%% file: egpaper.tex
\documentclass[10pt,twocolumn,letterpaper]{article}

\usepackage{iccv}
\usepackage{times}
\usepackage{epsfig}
\usepackage{graphicx}
\usepackage{amsmath}
\usepackage{amssymb}

\usepackage[accsupp]{axessibility}
\usepackage{subcaption}
\usepackage{authblk}
\newcommand{\alwod}{\texttt{ALWOD}\xspace}
\usepackage{verbatim}

\usepackage{booktabs}
\usepackage{mathtools}

\usepackage{soul}
\usepackage{xcolor}
\usepackage{xspace}
\usepackage{multirow}
\usepackage[pagebackref=true,breaklinks=true,letterpaper=true,colorlinks,bookmarks=false]{hyperref}
\usepackage[capitalize,nameinlink]
{cleveref}
\crefname{section}{Sec.}{Secs.}
\Crefname{section}{Section}{Sections}
\Crefname{table}{Table}{Tables}
\crefname{table}{Tab.}{Tabs.}
\crefname{figure}{Fig.}{Figs.}

\DeclareMathOperator*{\argmin}{argmin}
\DeclareMathOperator*{\argbmax}{argmax_B}

\iccvfinalcopy 

\def\iccvPaperID{9085} 

\ificcvfinal\fi

\begin{document}

\title{\alwod: Active Learning for Weakly-Supervised Object Detection}

\author[1]{Yuting Wang}
\author[2]{Velibor Ilic}
\author[1]{Jiatong Li}
\author[3,2]{Branislav Kisa\v{c}anin}
\author[1]{Vladimir Pavlovic}
\affil[1]{Rutgers University, NJ, USA}
\affil[2]{The Institute for Artificial Intelligence Research and Development of Serbia, Novi Sad, Serbia}
\affil[3]{Nvidia Corporation, TX, USA}
\affil[ ]{\normalsize{\texttt{yw632@rutgers.edu}, \texttt{velibor.ilic@ivi.ac.rs}, \texttt{jiatong.li@rutgers.edu}, \texttt{b.kisacanin@ieee.org}, \texttt{vladimir@cs.rutgers.edu}}
\vspace{-1.5em}
}
\maketitle
\ificcvfinal\thispagestyle{empty}\fi

\input{1_intro}
\input{2_related}

\input{3_method}

\input{4_exp}

\input{5_conclusion}

{\small
\bibliographystyle{ieee_fullname}
\bibliography{egbib}
}
\clearpage
\appendix
\input{6_appendix}

\end{document}

%% file: 1_intro.tex
\begin{abstract}
   Object detection (OD), a crucial vision task, remains challenged by the lack of large training datasets with precise object localization labels. In this work, we propose \alwod, a new framework that addresses this problem by fusing active learning (AL) with weakly and semi-supervised object detection paradigms.  Because the performance of AL critically depends on the model initialization, we propose a new auxiliary image generator strategy that utilizes an extremely small labeled set, coupled with a large weakly tagged set of images, as a warm-start for AL.
   We then propose a new AL acquisition function, another critical factor in AL success, that leverages the student-teacher OD pair disagreement and uncertainty to effectively propose the most informative images to annotate.   Finally, to complete the AL loop, we introduce a new labeling task delegated to human annotators, based on selection and correction of model-proposed detections, which is both rapid and effective in labeling the informative images.  We demonstrate, across several challenging benchmarks, that \alwod significantly narrows the gap between the ODs trained on few partially labeled but strategically selected image instances and those that rely on the fully-labeled data. Our code is publicly available on \url{https://github.com/seqam-lab/ALWOD}.  
\end{abstract}

\section{Introduction}
\label{sec:intro}

Object detection (OD) is a critical vision problem. To solve it, many fully-supervised object detection (FSOD) methods have been developed to build deep neural network architectures with high detection performance~\cite{ren2015faster,he2017mask,carion2020end} and fast inference~\cite{redmon2016you,redmon2017yolo9000}. Typically, these networks are trained on large fully-annotated (FA) data, which require humans to manually identify, with accurate bounding boxes and category labels, each object in an image. However, manual annotation of a large dataset is time-consuming~\cite{su2012crowdsourcing}, limiting the scalability of FSOD as the number of images, categories, and objects grows. To address this, many weakly-supervised object detection (WSOD) methods~\cite{bilen2016weakly,tang2017multiple, huang2020comprehensive,wangD2DF2WOD} have been developed. WSOD aims to reduce the object annotation cost by leveraging cheaper, weakly-annotated (WA) data, 
where an image instance is tagged according to the objects present in it, without the need to specify  bounding boxes for each object.

Existing WSOD methods often struggle to distinguish between object parts and objects, or between objects and groups of objects~\cite{ren2020instance}. The performance of WSOD methods lags behind that of FSODs since WSODs rely on weaker annotation signals. Recently proposed semi-supervised~\cite{radosavovic2018data,sohn2020simple,wang2022omni} and few-shot~\cite{biffi2020many,pan2019low} learning approaches demonstrated that a good trade-off between annotation effort and detection performance can be achieved by first fully annotating a set of random images, followed by training the detector on a combination of large WA and small FA data. Active learning (AL) methods~\cite{ yoo2019learning,choi2021active,yuan2021multiple,vo2022active} aim to further reduce the size of the FA sets using acquisition functions to select the most informative images for human labeling. AL methods can be either warm-start, which begin with a labeled set and iteratively select informative samples with feedback from the model, or cold-start, which select all informative samples at once without the need for an initial labeled set. Our work focuses on the warm-start setting.

\begin{figure}[ht!]
\centering
\vspace*{-0.2cm}
\includegraphics[trim={5.3cm 5.7cm 3.8cm 2.5cm},clip, scale=0.334]{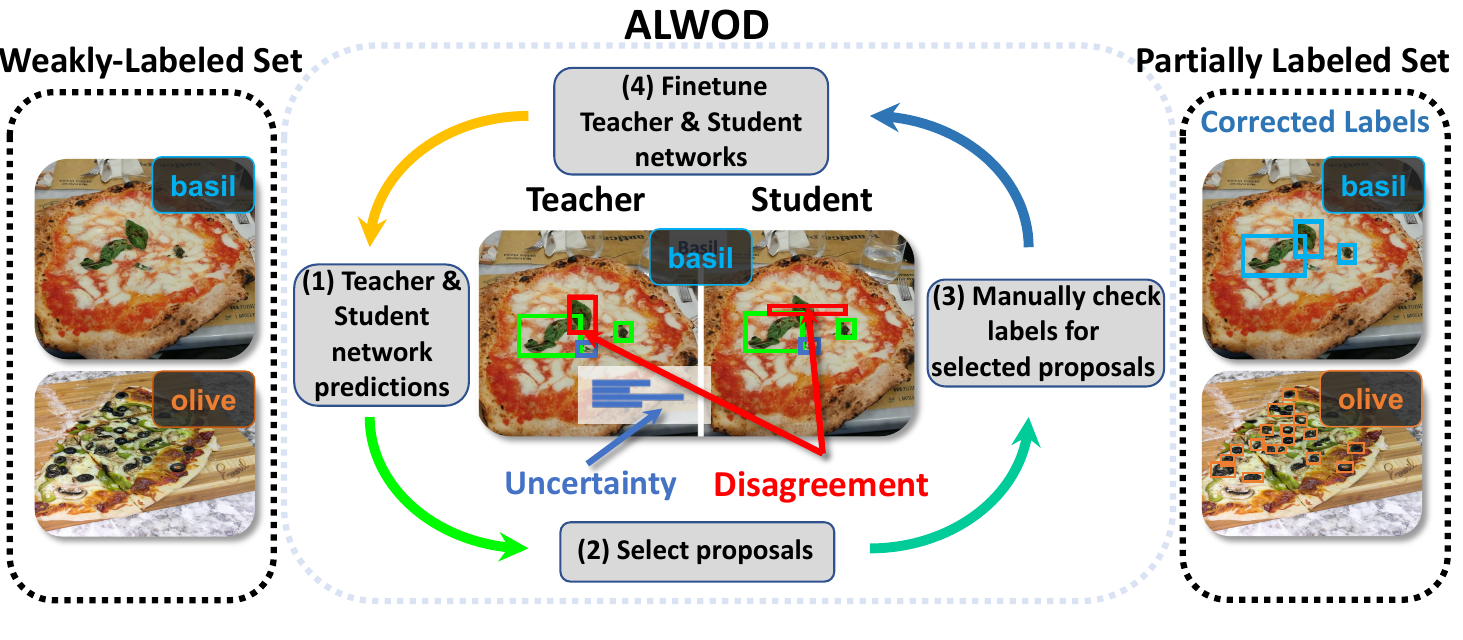}
\vspace*{-0.7cm}
\caption{
Overview of \alwod. The teacher and student networks are first semi-supervised trained on WA data and auxiliary FA training data generated by an image generator, and then fine-tuned in successive stages using a few new well-selected FA and large remaining WA data.
We propose a model disagreement and image uncertainty based acquisition function to select images, and build a new annotation tool to fully label selected images with decreased annotation workload compared to traditional AL labeling strategies.
}
\vspace*{-0.2cm}
\label{fig:gist}
\end{figure}

To reduce the annotation cost and maximize the detection performance, we introduce an {\em \textbf{A}ctive \textbf{L}earning for \textbf{W}eakly-Supervised \textbf{O}bject \textbf{D}etection} (\alwod) framework that combines semi-supervised learning with active learning by dynamically augmenting the semi-supervised set with a small set of actively selected and then fully annotated images, as illustrated in~\cref{fig:gist}. However, traditional warm-start AL methods commence with an FSOD model trained on a random set of FA data~\cite{choi2021active, elezi2022not}, typically hundreds or thousands of images, or a WSOD model trained on a large set of WA data~\cite{vo2022active}. While the former results in effective learning of OD models, it still requires a significant annotator effort at initialization; the latter strategy is less effective and necessitates more rounds of AL. To circumvent this and further reduce the annotation cost, we design an image generator that leverages few FA images to synthesize a large auxiliary FA 
FSOD training set. Together with the WA data, the two sets are used for semi-supervised pre-training of an OD. The auxiliary FA data can serve as a warm-start for existing AL approaches, which require initial FA data.

The effectiveness of AL is predicated on the selection of the most informative samples for human labeling. To that end, we propose a novel active learning function that naturally leverages a combination of the model disagreement between student-teacher based ODs~\cite{liu2021unbiased,wang2022omni} and the image uncertainty of the teacher network on the WA training data: images, where the network pairs disagree most and where the teacher exhibits high uncertainty, are passed on to human annotators for further labeling.
This strategy allows us to select informative samples for both the low and the high-performing classes, particularly on class-unbalanced data.

To further reduce the AL cost, we replace the typical annotator task of drawing bounding boxes and assigning to each an object class label with the task of \textit{correcting} the class labels of selected predicted bounding boxes, assigning bounding box quality scores, and removing falsely predicted bounding boxes. This type of annotation requires a significantly lower annotation effort while only slightly dampening the performance compared to the standard workload. 
The new FA and the remaining WA data will be used to fine-tune the student-teacher OD. This process is repeated until a desired annotation budget is met. As shown in~\cref{sec:exp}, our combined AL and WSOD approach can achieve performance on par with FSOD, while reducing the need to precisely annotate large image datasets.

\textbf{Our contributions} are three-fold: (1) We present a new 
framework \alwod to improve annotation efficiency
and annotation quality by combining active learning with weakly and semi-supervised object detection paradigms. 
(2) We introduce a new acquisition function that considers the model disagreement between student-teacher networks, coupled with image uncertainty. (3) We propose an auxiliary domain to 
warm up 
the learning process utilizing small labeled sets. Our experimental results demonstrate that \alwod achieves state-of-the-art performance across several benchmarks.

%% file: 2_related.tex
\section{Related Work}
\label{sec:related work}
\noindent\textbf{Weakly-Supervised Object Detection (WSOD).} WSOD methods generally aim to reduce detection annotation costs by exploiting only image-level annotations. Most existing methods mainly formulate WSOD as a multiple-instance learning (MIL) problem~\cite{bilen2016weakly,tang2017multiple, gao2019c,zeng2019wsod2,ren2020instance,huang2020comprehensive,huang2022w2n,wangD2DF2WOD}. Although
many promising results have been achieved by WSOD, they
are still not comparable to FSOD. Our work utilizes a small amount of FA data and a large amount of WA data based on active learning strategies and semi-supervised learning to achieve better performance.  

\noindent\textbf{Semi-Supervised Object Detection (SSOD).} SSOD work focuses on training an object detector with a combination of FA, WA, or un-annotated (UA) data. A traditional paradigm of SSOD is to construct a multi-stage self-training pipeline~\cite{rosenberg2005semi, sohn2020simple, wang2022omni}: (1) pre-train a model on FA data; (2) generate pseudo-labels on WA or UA data; (3) fine-tune the model on both FA and pseudo-labeled data; (4) repeat this process if needed. Some work~\cite{liu2021unbiased, wang2022omni,tang2021humble,liu2022unbiased,chen2022label,liu2023mixteacher} relies on a student-teacher framework, where the teacher network generates pseudo-labels for the student network. Our work is also based on a student-teacher framework, but extends it with active learning strategies for selecting most informative WA or UA data, which are fully annotated by humans.

\noindent\textbf{Active Learning for Object Detection (ALOD).} The traditional active learning strategies~\cite{ilic2022active,gal2016dropout,huang2021semi,guo2021semi,simeoni2021rethinking} are designed for classification tasks. Few active learning methods specifically focus on object detection~\cite{elezi2022not,vo2022active,choi2021active,wu2022entropy, yoo2019learning}, which is more challenging with complex instance distributions. ALOD methods aim to improve detection performance by selecting the most informative images to be fully annotated by humans. They define an acquisition function used to assign a single score representing the informativeness
of each weakly or unlabeled image. Most work~\cite{yoo2019learning,choi2021active,yuan2021multiple} considers the instance-based uncertainty as the acquisition signal. The work of \cite{kao2018localization} introduces localization tightness and localization stability metrics to quantitatively evaluate the localization uncertainty of an object detector. The work of \cite{choi2021active} proposes the aleatoric and epistemic uncertainty in both image and instance levels based on the Gaussian mixture model. Yoo \etal \cite{yoo2019learning} propose the active learning method with the loss prediction module to predict the loss of an input data point. Elezi \etal \cite{elezi2022not} introduce a class-agnostic active learning function based on the robustness of the network. In \cite{wu2022entropy}, the instance-level uncertainty and diversity are jointly considered in a bottom-up manner. Our work is related but different from the aforementioned methods. Similarly to these methods, we consider the image uncertainty as a part of the acquisition score. Unlike them, we also consider the model disagreement between student-teacher based ODs as a part of the  acquisition function which is even more reliable for both the low and the high performing classes.

A key to making ALOD approaches effective is to appropriately initialize the OD model. Choi \etal \cite{choi2021active} pre-train an FSOD model on a random set of fully-annotated data including 2,000 images, which requires a large annotation effort, and achieves 62.4\% AP50 on VOC2007 dataset. Vo \etal \cite{vo2022active} pre-train a WSOD model on weakly-annotated data with a small annotation effort, and achieves 47.7\% AP50 on VOC2007 dataset. 
It is essential to balance initial detection performance with annotation cost. In contrast to traditional approaches, our OD is pre-trained on a large fully-labeled auxiliary domain constructed with minimal effort from only a few (as low as 50) fully-annotated images and a large weakly-labeled domain in a semi-supervised manner. 

\noindent\textbf{Annotation workflow.} Traditional active learning approaches aim to query strong labels for data. However to reduce annotation costs, Desai \etal~\cite{desai2019adaptive} first query weak localization information by requiring humans to click the centers of objects. This point information is stronger than the image-level tag. Pardo \etal~\cite{pardo2021baod} first decide the type of annotation for each selected image then optimizes the detection model on the hybrid supervised dataset. To reduce annotation cost while maintaining high annotation quality, our method allows one to select an imprecise bounding box for each object, a stronger label signal than the marked points.

%% file: 3_method.tex
\section{Methodology}
\label{sec:method}
Our proposed object detection framework, \alwod, aims to address the lack of accurate object localization information in the real-world training data by formulating WSOD as a combination of semi-supervision and active learning. 

\subsection{Preliminaries and Problem Statement}

Consider an iterative, semi-supervised OD model learning setting where at each iteration $t=1,2,\ldots$ the model $M^t$ is learned from a 
dynamic combination of weakly and fully labeled data.  Let $W^t$, where $|W^t|=N-n$, and $F^t$, where $|F^t|=n$, denote the sets of indices of images in the training set $\mathcal{S}$ with the weak and full annotations, respectively, where $N$ is the number of images in $\mathcal{S}$.  An RGB image
$\mathbf{X}_j\in{\mathbb{R}}^{h{\times}w{\times}3}$, where $h$ and $w$ are its height and width, is said to be {\em fully annotated}, $j\in F^t$, if it is
associated with the label $\mathbf{Y}^f_{j} = \{(\mathbf{b}_{k},c_{k},p_{k})\}_{k=1}^{n^f}$ for each of the $n^f$ objects present (labeled) in that image. The label consists of $\mathbf{b}_{k}\in{\mathbb{R}}^{4}$, the $k$-th object's localization bounding box defined by $(x_\mathrm{min},y_\mathrm{min},x_\mathrm{max},y_\mathrm{max})$ that specifies its top-left corner $(x_\mathrm{min},y_\mathrm{min})$ and its bottom-right corner $(x_\mathrm{max},y_\mathrm{max})$.  The label also contains the class label $c_{k} \in \{1,\ldots,C \}$, where C is the number of object
categories, and the bounding box quality score $p_{k} \in \{ 1, 0\}$\footnote{The score corresponds to a subjective (annotator) notion of whether the (predicted) bounding box is precise, $p_k = 1: \mathrm{IoU}\geq 0.9$, or imprecise, $p_k = 0: 0.5<\mathrm{IoU}<0.9$. See~\cref{sec:annotation}.}.
The same image is said to be {\em weakly annotated}, $j\in W^t$, if the image label contains only the classes of objects present in that image but not the objects' locations, \ie , $\mathbf{Y}^w_{j} = \{c_{k}\}_{k=1}^{n^w}$, where $1 \leq n^w \leq C$ is the number of object classes in that image.  We denote this ``version'' of the dataset
$\mathcal{S}^t \coloneqq \mathcal{S}(W^t,F^t)$.

\begin{figure}[ht!]
\centering
\vspace{-0.2cm}
\includegraphics[trim={0cm 1.8cm 0 3.8cm},clip, width=\columnwidth]{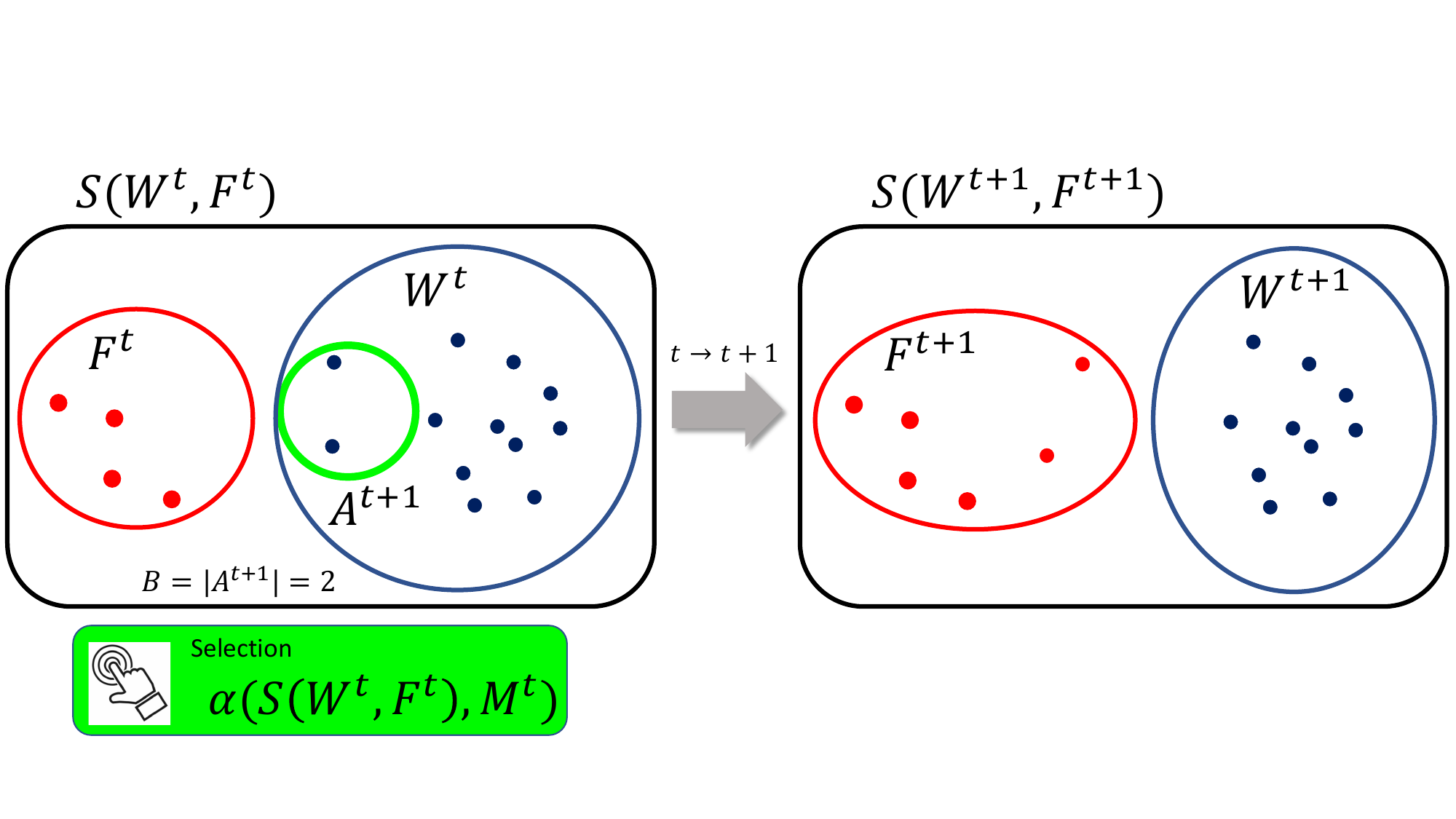}
\caption{Training set notation for two steps of active learning for object detectors $M^t$. 
}
\vspace{-0.2cm}
\label{fig:update_data}
\end{figure}

After model $M^t$ is learned at cycle $t$ from $\mathcal{S}(W^t,F^t)$, an active learning acquisition function $\alpha(\mathcal{S}(W^t,F^t), M^t)$ will select a set of $B$ weakly annotated images with indices $A^{t+1} \subseteq W^t$, $|A^{t+1}|=B$, which will be passed on to a human annotator to label. The selection will be based on an assessment of model $M^t$'s performance on $\mathcal{S}^t$ according to existing full and weak labels over $F^t \cup W^t$. In this fashion, we will arrive at an updated ``version'' $\mathcal{S}(W^{t+1},F^{t+1})$, where $F^{t+1} = F^{t} \cup A^{t+1}$ and $W^{t+1} = W^{t} \setminus A^{t+1}$, which will have $B$ more fully annotated images $|F^{t+1}| = |F^{t}| + B$ and $B$ fewer weakly annotated images $|W^{t+1}| = |W^{t}| - B$ than the previous $\mathcal{S}^t$.
This process is illustrated in \cref{fig:update_data}.

Our goal here is to design the acquisition function $\alpha()$ and the semi-supervised learning approach for learning the OD models $M^t$ to minimize the annotator workload proxied through the annotation budget $T\cdot B$, where $T$ is the total number of cycles of active learning, while maximizing the final model detection accuracy. As a part of this process, we also aim to design the initialization procedure for model $M^0$.  The following sections describe our proposed approach to achieving these goals.

\subsection{Object Detection with Student-Teacher Networks}

\begin{figure*}[ht!]
\centering
\vspace*{-0.5\baselineskip}
\includegraphics[trim={3cm 3cm 7cm 4cm},clip, scale=0.6]{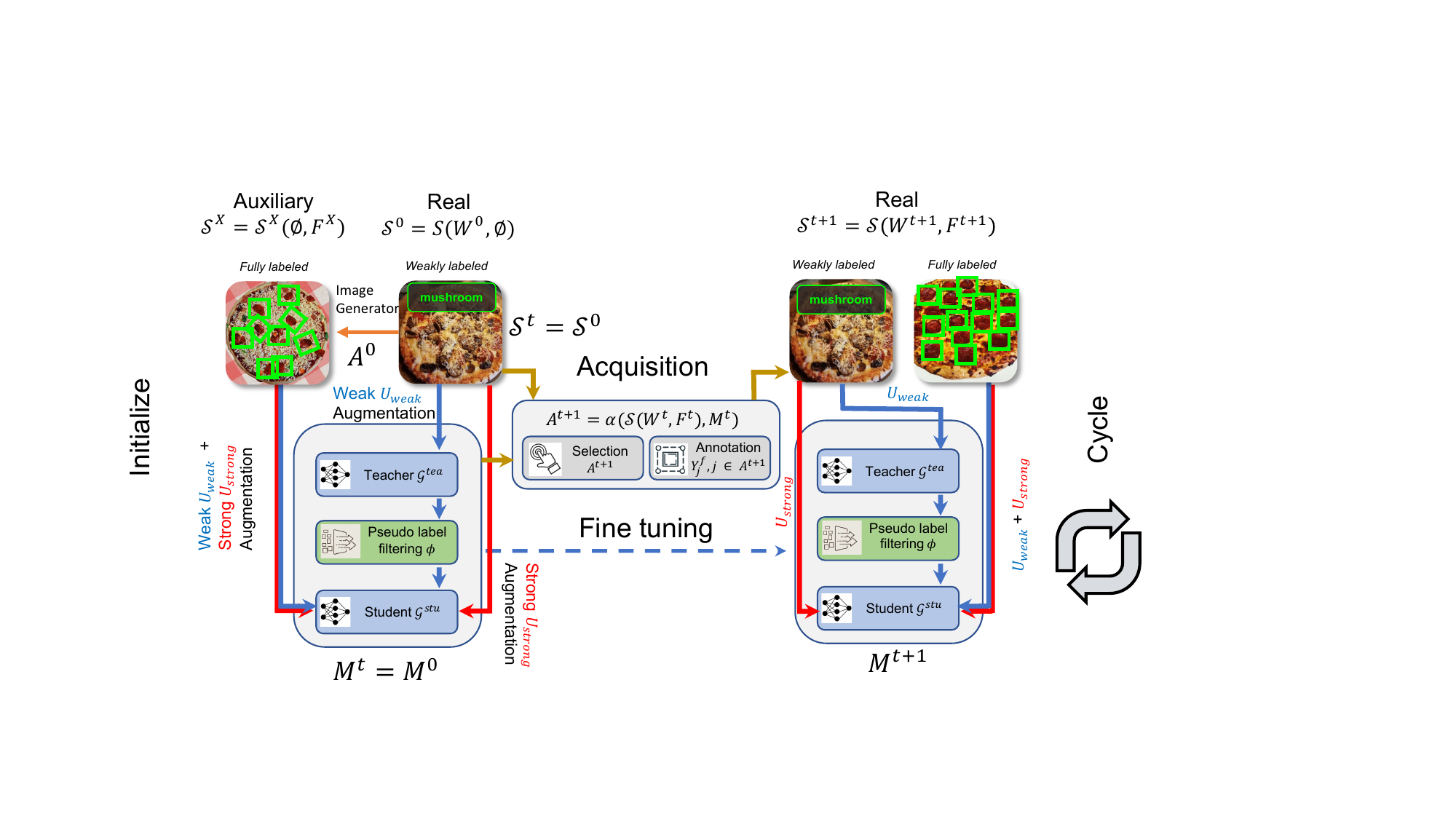}
\caption{Architecture of \alwod. The framework couples student-teacher WSOD and SSOD with active learning. The initial OD $M^0$ is pre-trained on $\mathcal{S}^0 \cup \mathcal{S}^X$ in a semi-supervised manner. The auxiliary $\mathcal{S}^X$ is created using an image generator and a small, fully-annotated $A^0$. At each cycle $t > 0$, we select $B$ images using a novel acquisition function $\alpha(\cdot)$, which are passed on to annotators for full labeling using our annotation tool. The $M^{t}$ is fine-tuned on the updated $\mathcal{S}^{t+1}$. 
}
\label{fig:architecture}
\vspace*{-0.5\baselineskip}
\end{figure*}

\cref{fig:architecture} gives an overview of our framework. Motivated  by the recent success of student-teacher networks for semi-supervised learning~\cite{liu2021unbiased,wang2022omni} and transformer-based object detector~\cite{carion2020end,roh2021sparse}, our semi-supervised object detector is composed of a student detection network $\mathcal{G}^{stu}(\mathbf{X}|\theta_{stu})$ and a teacher detection network  $\mathcal{G}^{tea}(\mathbf{X}|\theta_{tea})$. Both $\mathcal{G}^{stu}$ and $\mathcal{G}^{tea}$ are transformer-based object detectors or Faster-RCNN~\cite{ren2015faster}\footnote{Since transformer-based object detectors are stronger than Faster-RCNN, we focus on transformer-based object detectors.}.  Thus, $M^t \coloneqq \{ \mathcal{G}^{stu}(\cdot|\theta_{stu}^t), \mathcal{G}^{tea}(\cdot|\theta_{tea}^t) \}$, where $\theta_{stu}^t$ and $\theta_{tea}^t$ are the models' parameters at stage $t$ of active learning.

\subsubsection{Initial Learning ($t=0$)}\label{sec:init_learn}

\noindent\textbf{Assumptions:} Our learning of the OD model begins with an assumption that the entire initial set of real-world images $\mathcal{S}$ is only tagged with weak labels, \ie, $W^0 = \{1,\ldots,N\},F^0 = \emptyset$, which we will denote $\mathcal{S}^0\coloneqq \mathcal{S}(W^0,F^0)$.  This, while realistic, is an extremely challenging assumption and will lead to poor initial OD models and high annotation workload, \ie, large $B$.  To address this challenge, we make the second assumption that one can ``cheaply'' construct a large \textit{auxiliary} ``synthetic'' fully-labeled dataset, which we denote $\mathcal{S}^X \coloneqq \mathcal{S}^{X}(W^{X}= \emptyset,F^{X} = \{1,\ldots,N_{X}\})$.  We will use $\mathcal{S}^0 \cup \mathcal{S}^X$ to initialize our student-teacher OD model\footnote{Note that $\mathcal{S}^X$ is only used in the initial cycle $t=0$.}.

\noindent\textbf{Auxiliary set:} Performance of OD learning critically depends on model initialization.
It is thus essential to pre-train an FSOD model that achieves good detection performance with the least annotation cost. To that end, $\mathcal{S}^X$ is created by combining real-world background images with synthetically ``pasted'' but otherwise real-in-appearance foreground objects. These realistic foreground objects are cropped from the FA images in $A^0$, which are randomly selected from $S^0$. The annotation cost of this large fully-annotated auxiliary set is identical to that of the $B$ images in $A^0$. However, as demonstrated in \cref{sec:ablation}, the role of $\mathcal{S}^X$ is essential, when combined with  $\mathcal{S}^0$, to learn an effective ALOD.

\noindent\textbf{Burn-in:} $\mathcal{G}^{stu}$ is first trained only on fully-labeled data $\mathcal{S}^X$.
The teacher $\mathcal{G}^{tea}$ is initialized by duplicating the burn-in student model, $\theta_{tea} = \theta_{stu}$. 

\noindent\textbf{Student-teacher learning:} We build upon the student-teacher learning paradigm of~\cite{wang2022omni}. Therein, two types of augmentation (strong and weak) are used to regularize learning of the student-teacher network pair.  Here, we train this pair using $\mathcal{S}^0 \cup \mathcal{S}^X$.  

Specifically, we apply both weak $\mathrm{U}_{weak}(\cdot)$ and strong $\mathrm{U}_{strong}(\cdot)$ augmentation to the data, expanding the fully-labeled initial set to $\mathrm{U}_{weak}(\mathcal{S}^X) \cup \mathrm{U}_{strong}(\mathcal{S}^X)$ and the weakly labeled real-world set to $\mathrm{U}_{weak}(\mathcal{S}^0) \cup \mathrm{U}_{strong}(\mathcal{S}^0)$.  We then train the student network using $\mathrm{U}_{weak}(\mathcal{S}^X) \cup \mathrm{U}_{strong}(\mathcal{S}^X)$ as well as the pseudo-labeled $\mathrm{U}_{strong}(S^0)$, with labels proposed by $\mathcal{G}^{tea}(\mathrm{U}_{weak}(S^0))$\footnote{
We use the following notation for brevity: $\mathrm{U}_a(\mathcal{S})$ means that the image component $\mathbf{X}$ of $(\mathbf{X},\mathbf{Y}) \in \mathcal{S}$ is transformed by the augmentation operator $\mathrm{U}_a(\cdot)$, i.e., $\mathrm{U}_a(\mathcal{S}) \coloneqq \{ 
 (\mathbf{X}',\mathbf{Y}') : \mathbf{X}' = \mathrm{U}_a(\mathbf{X}), \mathbf{Y}' = \mathrm{U}_a(\mathbf{Y}), \forall (\mathbf{X},\mathbf{Y}) \in \mathcal{S} \}$. Augmentation of labels is applied as necessary to enforce label coherence, e.g., when $\mathrm{U}_a(\cdot)$ is L-R flip, the object class labels are maintained while the object locations are "flipped".
 Similarly, $\mathcal{G}^m(\mathcal{S})$ stands for the set constructed by replacing the label component $\mathbf{Y}$ of the $(\mathbf{X},\mathbf{Y}) \in \mathcal{S}$ pair using the predictive model $\mathcal{G}^m(\cdot)$.  I.e., 
$\mathcal{G}^m(\mathcal{S}) \coloneqq \{ 
 (\mathbf{X}',\mathbf{Y}') : \mathbf{X}' = \mathbf{X}, \mathbf{Y}' = \mathcal{G}^m(\mathbf{X}), \forall (\mathbf{X},\mathbf{Y}) \in \mathcal{S} \}$.
 }  and filtered by $\phi(\cdot)$:
\begin{equation}
    \theta_{stu}^0 \leftarrow \min \sum_{(\mathbf{X},\mathbf{Y}) \in \mathcal{T}^0 } \mathcal{L}(\mathcal{G}^{stu}(\mathbf{X}|\theta_{stu}),\mathbf{Y}),
    \label{eq:learnstu0}
\end{equation}
where
\begin{equation}
    \mathcal{T}^0 = \mathrm{U}_{weak}(\mathcal{S}^X) \cup \mathrm{U}_{strong}(\mathcal{S}^X) \cup \phi(\mathcal{G}^{tea}(\mathrm{U}_{weak}(S^0))),
    \label{eq:learnstu_train0}
\end{equation}
and $\mathcal{L}$ is the classification and bounding box regression loss used in transformer-based detectors~\cite{carion2020end,roh2021sparse}. Our pseudo-labeling filtering approach $\phi(\cdot)$ uses the bounding box quality score and the bounding box tag while~\cite{wang2022omni} does not. For details of $\phi(\cdot)$, please refer to \cref{sec:PL}. 

The teacher network is updated by the exponential moving average (EMA) from the student network~\cite{tarvainen2017weight},
\begin{equation}
    \theta_{tea} \leftarrow{q\theta_{tea} + (1-q)\theta_{stu}},
\end{equation}
where $q\!\in\!(0,1)$, set empirically to $q\! =\! 0.9996$. This EMA updated teacher can be seen as a temporal ensemble of student models along the training trajectories~\cite{wang2022omni}. After training $M^0$, we set $F^0 \leftarrow F^0 \cup A^0$ and $W^0 \leftarrow W^0 \setminus A^0$. 

In \cref{sec:activelearn}, we introduce a novel way of exploiting the model disagreement and the image uncertainty between the student and the teacher networks to identify informative samples, which are subsequently used to minimize the human annotation effort. 

\subsubsection{Active Learning Cycle ($t>0$)}
At each cycle $t>0$, we select $B$ images from the current weakly-annotated data $W^{t}$ using our acquisition function $A^{t+1} = \alpha(\mathcal{S}(W^t,F^t), M^t)$. The human annotator will annotate images in this acquisition set $A^{t+1}$ with bounding boxes, object class labels, and bounding box quality scores, helped by predictions generated from $M^{t}$. This results in a new set of full labels for images in this set, $\mathbf{Y}_j^f, j \in A^{t+1}$, replacing their existing weak labels.  In this manner, we have created a new training set $\mathcal{S}^{t+1} = \mathcal{S}(W^{t+1},F^{t+1})$, increasing by $B$ images the fully labeled image set $F^t$ to $F^{t+1}$. Since the new fully-annotated images may include imprecise bounding boxes $b_k$, where $p_k=0$, we propose a new pseudo-labeling filtering method embodied in $\phi(\cdot)$. For an imprecise bounding box $b_k$, using the strategy of~\cite{wang2022omni}, we find the best matched predicted bounding boxes $\hat b_k$. We then generate a final pseudo-labeled bounding box by interpolating the coordinates of the imprecise and matched boxes. The precise bounding boxes coincide with the final pseudo-labeled boxes.
$\mathcal{S}^{t+1}$ will be used to fine-tune the student-teacher pair $(\theta_{stu}^t,\theta_{tea}^t)$, using the new pseudo-labeling filtering,  resulting in an updated OD $M^{t+1}$.  This cycle will repeat $T$ times until the annotation budget $T\cdot B$ is met or the model converges in validation loss. The detection performance is evaluated with the teacher network.

The key to making this active learning process effective is to define an optimal acquisition function $\alpha(\cdot)$ and the annotation workflow. In the next section, we propose a novel approach to designing such a function together with a new annotation procedure.

\subsection{Active Learning Strategies}\label{sec:activelearn}\label{sec:acq_score}
We aim to select the most informative training samples $A^{t+1}$ from $\mathcal{S}^t$, with a small annotation budget $B$, that will lead to the largest reduction in loss (improvement in detection performance), based on the trained student and teacher networks. For this, we consider the following two signals: (a) the disagreement between the teacher-student network pair, and (b) the uncertainty of predictions on each image. We first define the scores for each signal, then fuse them to arrive at the final acquisition function. 

\noindent\textbf{Model Disagreement.}
The EMA updated teacher behaves as a stochastic average of consecutive student models~\cite{tarvainen2017mean}.
In an ideal case, the student network's predictions would be consistent with the teacher's predictions. This naturally leads to using the disagreement of predictions between student and teacher networks as one of the acquisition signals to create $A^{t+1}$. 
Specifically, we define the model disagreement acquisition score $\beta_{MD}(\cdot)$ on image $\mathbf{X}$ as
\begin{equation}
    \beta_{MD}(\mathbf{X} | \mathcal{S}^t, M^t) \coloneqq
      1 - \frac{\sum_{c} AP_{c}(\mathcal{G}^{stu}(\mathbf{X}) | \mathcal{G}^{tea}(\mathbf{X}) )} { n^w}, 
    \label{eq:acqs_mi}
\end{equation}
where $c \in \{c_{k}\}_{k=1}^{n^w}$ are the known classes of objects present in $\mathbf{X}$.  This scores an image according to the value of the average precision per-class score of the student model predictions when treating the teacher model predictions as the ``ground truth''. The higher the score, the less agreement the models have on image $\mathbf{X}$, indicating that the image may be a plausible candidate for manual annotation.

\noindent\textbf{Image Uncertainty.}
Another traditional signal that points toward the need to manually annotate an image is the class prediction uncertainty. Given the uncertainty for each prediction in an image,  we define the uncertainty of the image by aggregating the score over all predicted objects. Here, we define the image uncertainty score as the maximum entropy $\beta_{IU}(\cdot)$ on image $\mathbf{X}$ for the $n^f$ objects predicted by the teacher network:
\begin{equation}
    \beta_{IU}(\mathbf{X} | \mathcal{S}^t, M^t) \coloneqq
     \max_{k=1}^{n^f}{H(c_k|\theta_{tea})},
\end{equation}
where $H(c_k|\theta_{tea})$ represents the entropy over the distribution $c_k$ of predictions generated from the teacher network.
The higher the entropy, the more uncertain the model is about its prediction, indicating that the image may need to be manually annotated.

\noindent\textbf{Acquisition Function.}
We propose the final acquisition function for each image, which fuses the model disagreement and the image
uncertainty signals:
\begin{multline}
    \alpha_{\Sigma}(\mathcal{S}(W^t,F^t), M^t) \coloneqq \\
    \argbmax_{j \in W^t}  \beta_{MD}(\mathbf{X}_j | \mathcal{S}^t, M^t) + \beta_{IU}(\mathbf{X}_j | \mathcal{S}^t, M^t), 
    \label{eq:acqfunsum}
\end{multline}
or
\begin{multline}
    \alpha_{\Pi}(\mathcal{S}(W^t,F^t), M^t) \coloneqq \\
    \argbmax_{j \in W^t}  \beta_{MD}(\mathbf{X}_j | \mathcal{S}^t, M^t) \cdot \beta_{IU}(\mathbf{X}_j | \mathcal{S}^t, M^t), 
    \label{eq:acqfunpro}
\end{multline}
which selects $B$ images from the weakly labeled set $W^t$ with the highest values of the total score. We empirically find that taking the product of the model disagreement and the image uncertainty scores performs better than taking the sum\footnote{We use $\alwod_{\Sigma}$ to denote the model based on $\alpha_{\Sigma}(\mathcal{S}(W^t,F^t), M^t)$ and $\alwod_{\Pi}$ to denote the model that uses $\alpha_{\Pi}(\mathcal{S}(W^t,F^t), M^t)$ active learning strategy. When not specified, \alwod refers to $\alwod_{\Pi}$.}. Intuitively, for each cycle $t$ of active learning, we will be selecting those images where the student-teacher models disagree the most and the teacher model predictions are the most uncertain.

\subsection{Annotation Procedure and Tool}\label{sec:annotation}

\begin{figure}[t]
\centering
 \includegraphics[width=\columnwidth]{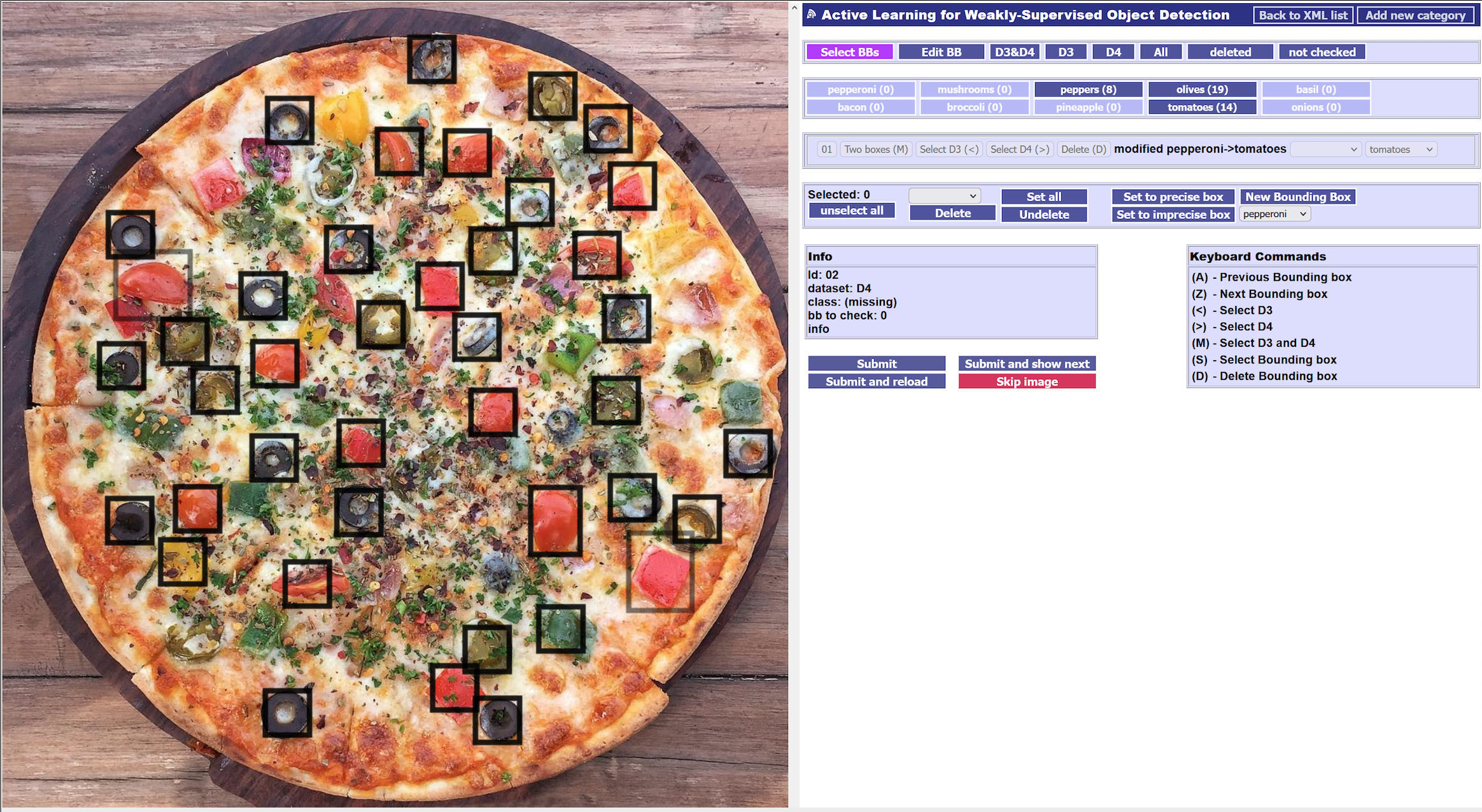} 
 \caption{Web tool for manual checking, editing bounding boxes, and correcting classes.
}
\label{fig:interface}
\vspace*{-1\baselineskip}
\end{figure}

In traditional OD annotation,  for each image humans are asked to draw tight bounding boxes around the objects to be detected and then select categories for each bounding box, an expensive and tedious task~\cite{dollar2009pedestrian,everingham2010pascal,gupta2019lvis}. For instance, \cite{su2012crowdsourcing} reports an average drawing box annotation time of 34.5 seconds (25.5 seconds for drawing one box and 9.0 seconds for verifying quality). Extreme clicking~\cite{papadopoulos2017extreme} relaxes the task of clicking on four extreme points of the object and requires about 7 seconds. Considering the trade-off between annotation efficiency and annotation quality, we develop an annotation tool that leverages our acquisition scores from \cref{sec:acq_score} to improve the efficacy of labeling. An example of the tool's frontend is shown in~\cref{fig:interface}.

Specifically, all images from $A^{t+1}$ are presented to the user in an ordered list according to \eqref{eq:acqfunpro}. Each image contains a large number of predicted bounding boxes with predicted class labels generated from both student and teacher networks. After applying non-maximum suppression and confidence threshold, some proposals with bounding boxes and class labels are given to the annotators. The annotators are asked to: (1) select the bounding boxes from proposals that overlap with true objects ($>50\%$ IoU) and include at least one of the four extreme points (top, bottom, left-most, right-most), (2) correct the bounding box categories, and (3) assess the bounding box quality.
The remaining unselected bounding boxes are removed.
If there is no bounding box over an actual object, the annotators directly draw a tight bounding box and select the object label. 

Compared to~\cite{su2012crowdsourcing,papadopoulos2017extreme}, our annotation process is significantly cheaper: on average, it can be completed in only 2 seconds, a reduction of over $70\%$ over~\cite{papadopoulos2017extreme}.  This results in a significantly decreased overall workload, as demonstrated in \cref{sec:exp}. For details of the annotation procedure, please refer to \cref{sec:annotation_sup}.

%% file: 4_exp.tex
\section{Experimental Results}
\label{sec:exp}
\noindent\textbf{Datasets and Evaluation.}
We evaluate our method on three object
detection benchmarks, VOC2007~\cite{everingham2010pascal}, COCO2014~\cite{lin2014microsoft}, and RealPizza10~\cite{wangD2DF2WOD}.
Following previous works~\cite{vo2022active,wangD2DF2WOD}, we use the trainval split of VOC2007 for training and the test split for evaluation, respectively containing 5,011 and 4,952 images. On COCO2014, we train detectors with the train split (82,783 images) and evaluate on the validation split (40,504 images). We use trainval split of RealPizza10 with 5,029 images for training and 552 test split images for testing. 
On RealPizza10, COCO2014, and VOC2007 datasets, each image contains 19.1, 7.7, and 2.5 instances on average, respectively. To evaluate the detection performance, we use the average
precision metrics AP50 and AP, computed with the IoU threshold of $\tau = 0.5$ and the threshold set $\tau \in \{ 0.5, 0.55, \ldots, 0.95 \}$, respectively. A predicted box is treated as a positive example when the IoU between the ground truth bounding box and the predicted object box exceeds $\tau$. 

\begin{figure}[t]
\centering
\includegraphics[trim={0cm 0cm 0cm 0},clip, scale=0.24]{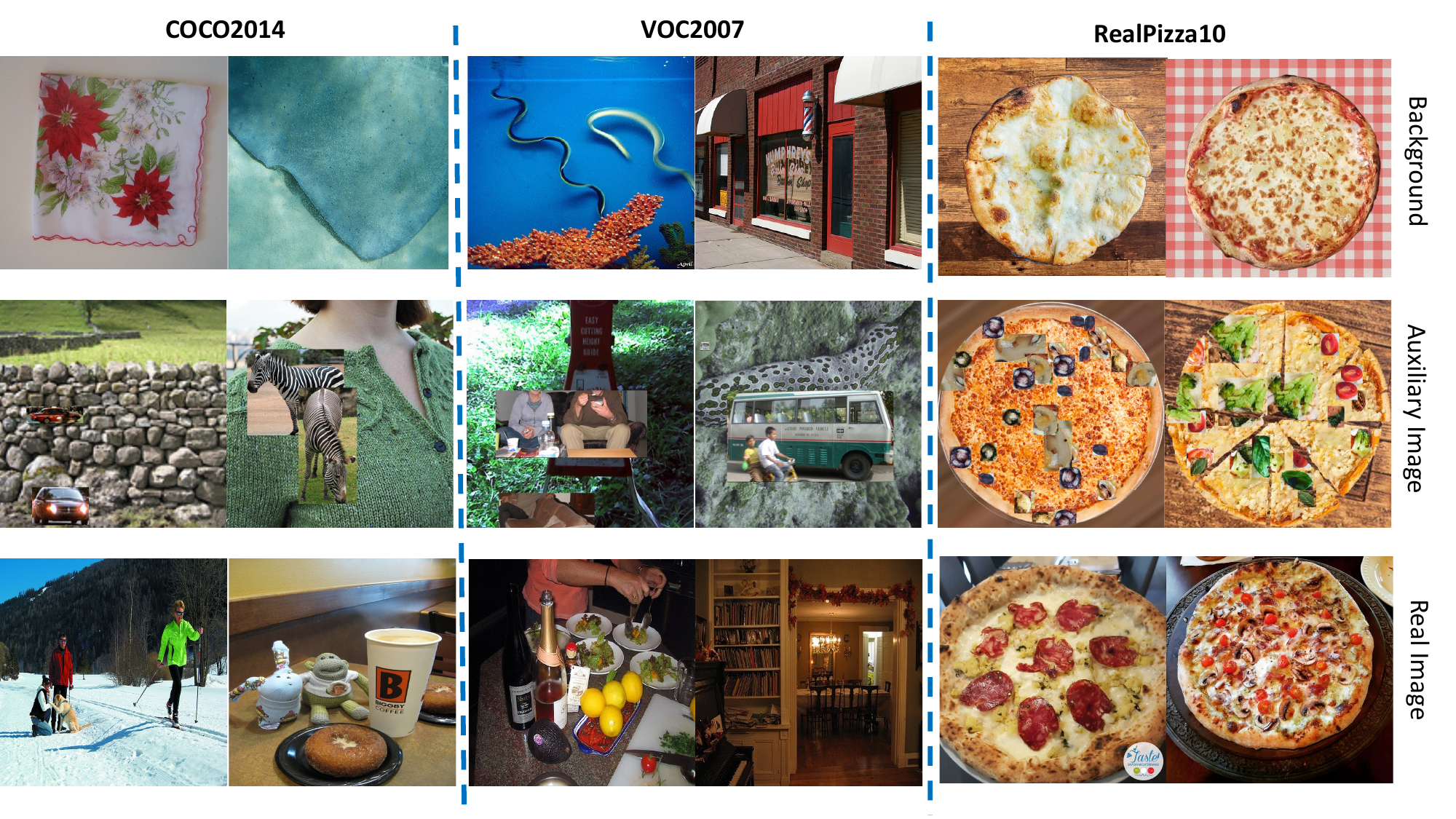}
\caption{Examples of background images, auxiliary images, and real images on COCO2014, VOC2007, and RealPizza10 datasets.
}
\label{fig:image}
\vspace*{-1.5\baselineskip}
\end{figure}

\noindent\textbf{Auxiliary Image Generator.}
The image generator creates auxiliary images in $\mathcal{S}^X$ by composing background images and object templates, as illustrated in~\cref{fig:image}. The object templates are created by cropping the object instances of fully-annotated images in $A^{0}$. Image augmentations such as rotation and scaling are performed on these templates, after which they are placed at random locations over the background images by employing a copy-paste augmentation technique sourced in~\cite{yun2019cutmix,ghiasi2021simple}. For details of the image generator, please refer to \cref{sec:auxilary_domain}. 

\noindent\textbf{Implementation Details.}
The VGG16~\cite{simonyan2014very}, ResNet50~\cite{he2016deep}, and Swin-T~\cite{liu2021swin} models pre-trained on ImageNet~\cite{russakovsky2015imagenet} were used as backbones. We adopt Sparse DETR~\cite{roh2021sparse} or Faster-RCNN~\cite{ren2015faster} for both $\mathcal{G}^{stu}$ and $\mathcal{G}^{tea}$. The confidence threshold for pseudo-labeling is $0.7$.
Following~\cite{liu2021unbiased}, for strong augmentations, we apply random color jittering, grayscale, Gaussian blur, and cutout patches. For weak augmentations, only random horizontal flipping is used. We report the results of ``$L\%$" experiments, where about $L\%$ of the training weakly-labeled set are selected to be fully annotated.
The minimal image height and width are set to 800 pixels. On VOC2007 and RealPizza10 datasets, $N_X = 2 N$, and on COCO2014, $N_X = 0.5N$.

\begin{table*}[htb]
    \centering
    \caption{Results (AP50 and AP in \%) for different methods in different settings on VOC2007, COCO2014, and RealPizza10. In FSOD and WSOD settings, each baseline considers the same fraction of the FA images across datasets. In SSOD and ALOD settings, different approaches consider different fractions of FA images in different benchmarks, \eg, \alwod and BiB consider 5\% of FA data on VOC2007, 1\% on COCO2014, and 5\% on RealPizza10, respectively, denoted as $5\% / 1\% / 5\%$. Red figures denote the best-performing non-FSOD method, followed by the second-best in blue. * denotes reproduced results.
    } 
    \label{tab:voc_coco_baseline}
    \resizebox{0.8\textwidth}{!}{
    \begin{tabular}{ccccc c c c}
    \toprule
    \multirow{2}{*}{Setting} & \multirow{2}{*}{$n/N$} & \multirow{2}{*}{Backbone (Detector)}& \multirow{2}{*}{Method} & VOC2007 & \multicolumn{2}{c}{COCO2014} & RealPizza10\\
    & & & & AP50 & AP50 & AP & AP50\\
    \midrule
    \multirow{5}{*}{FSOD} & \multirow{5}{*}{$100\%$}
    & VGG16 & Faster-RCNN~\cite{ren2015faster} & 69.9 & 42.1 & 20.5 & 39.1 \\
    & & VGG16 & SSD~\cite{liu2016ssd} & 68.0 & 42.1 & 24.1 & 32.8\\
    & & VGG16 & Sparse DETR~\cite{roh2021sparse} & 72.1 & 58.1 & 33.4 & 41.2\\
    & & ResNet50 & Faster-RCNN~\cite{ren2015faster} & 74.1 & 59.1& 38.4& 40.2\\
    & & ResNet50 & Sparse DETR~\cite{roh2021sparse} & 88.4 & 65.8 & 45.5 & 42.7 \\
    & & Swin-T & Sparse DETR~\cite{roh2021sparse} & 90.2 & 69.2 & 48.2 & 43.8\\   
    \midrule
    \multirow{8}{*}{WSOD} & \multirow{9}{*}{$0\%$}
    & \multirow{9}{*}{VGG16}& WSDDN~\cite{bilen2016weakly}& 34.8 & 11.5 & - & -\\
    & & & OICR~\cite{tang2017multiple} & 41.2 & - & - & 4.7\\
    & & & C-MIDN~\cite{gao2019c} & 52.6 & 21.4 & 9.6 & -\\
    & & & WSOD2~\cite{zeng2019wsod2} & 53.6 & 22.7 & 10.8 & -\\
    & & & MIST(+Reg)~\cite{ren2020instance} & 54.9 & 24.3 & 11.4 & -\\
    & & & CASD~\cite{huang2020comprehensive} & 56.8 & 26.4 & 12.8 & 12.9\\
    & & & W2N~\cite{huang2022w2n} & 65.4 & - & - & -\\
    & & & D2DF2WOD~\cite{wangD2DF2WOD} & 66.9 & - & - & 25.1 \\
    \midrule
    \multirow{4}{*}{SSOD} & $10\%$
    & VGG16
    & BCNet~\cite{pan2019low} & 61.8 & 38.3 & 22.9 & 35.4\\
    & $10\%$ & VGG16 & OAM~\cite{biffi2020many} & 63.3 & - &  - & - \\
    & $5\% / 1\% / 5\%$
    & ResNet50 (Faster-RCNN) & Unbiased teacher v2~\cite{liu2022unbiased} &  61.2 & 37.5 & 22.3 & 32.8 \\
    & $5\% / 1\% / 5\%$ & ResNet50 (Faster-RCNN) & Label Matching~\cite{chen2022label} & 61.7 & 37.7 & 22.4 & 34.1 \\
    \midrule
    \multirow{10}{*}{ALOD} & {$10\%$} & VGG16 (Faster-RCNN) & BAOD~\cite{pardo2021baod} & 50.9 & - & - & - \\
    & {$80\% / 8\% / 80\%$} & VGG16(SSD) & SSDGMM~\cite{choi2021active}* & 62.1 & 28.8 & 15.3 & 23.4 \\
    & {$80\% / 12.5\% / 80\%$}& VGG16 (SSD)& NALAE~\cite{elezi2022not}* & 67.7  & 25.5 & 11.9  & 32.3\\
    \cline{2-8} 
    & \multirow{8}{*}{$5\% / 1\% / 5\%$}
    & VGG16 (SSD) & SSDGMM~\cite{choi2021active} & 28.8 & 8.7 & 4.3 & 16.4 \\
    & & VGG16 (SSD) & NALAE~\cite{elezi2022not} & 36.3 & 8.8 & 3.4 & 22.2 \\
    & & Resnet50 (Faster-RCNN) & Active Teacher~\cite{mi2022active} & 49.7 &33.5 & 18.0& 30.8 \\
    & & VGG16 (MIST~\cite{ren2020instance}) & BiB~\cite{vo2022active}* & 60.6 & 33.2 & 16.5 &15.7 \\
    & & ResNet50 (Faster-RCNN) & \alwod & 69.1 & 39.8 & 24.5 & 38.0 \\
    & & VGG16 (Sparse DETR) & \alwod & 68.0 & 38.4 & 23.7 & 37.9\\
    & & ResNet50 (Sparse DETR) & \alwod & \color{blue}{70.5} & \color{blue}{41.8} & \color{blue}{26.0} &\color{blue}{39.3} \\
    & & Swin-T (Sparse DETR) & \alwod & \color{red}71.7 & \color{red}{42.5} & \color{red}{27.2} & \color{red}{40.2} \\

    \end{tabular}
    }
\end{table*}

\begin{table*}[hbt]
    \centering
    \caption{Effectiveness of the auxiliary domain $\mathcal{S}^X$ on RealPizza10 and VOC2007. For RealPizza10, the set cardinalities are: $|\mathcal{A}^0| = 50$, $|\mathcal{S}^X| = 10,058$, and $|\mathcal{S}^0| = 5,029$. For VOC2007, the set cardinalities are: $|\mathcal{A}^0| = 50$, $|\mathcal{S}^X| = 10,022$, and $|\mathcal{S}^0| = 5,011$. To focus on the auxiliary domain, we reproduce all numbers by applying each training set to each method.}
    \label{tab:auxiliary_ab}
    \resizebox{0.8\textwidth}{!}{
    \begin{tabular}{ccc|ccc|ccc}
    \toprule
    \multirow{2}{*}{Backbone} & \multirow{2}{*}{Setting} & \multirow{2}{*}{Training set} & \multicolumn{6}{c}{AP50 (\%)} \\
    \cline{4-9}
    & & & \multicolumn{3}{c|}{RealPizza} & \multicolumn{3}{c}{VOC2007} \\
    & & & SSDGMM & BiB & \alwod & SSDGMM & BiB & \alwod \\
    \midrule
    \multirow{5}{*}{VGG16}  
    & \multirow{5}{*}{ALOD} & 
    $\mathcal{A}^0$ & 10.7 & - & 4.3 & 14.4 & - & 3.3\\
    \cline{3-9}
    & & $\mathcal{S}^0$ & - &11.8 & - & - & 47.7 & - \\
    \cline{3-9}
    & & $\mathcal{S}^X$ & 10.8 & - & 7.0 & 14.6 & - & 39.2 \\
    \cline{3-9}
    & & $\mathcal{A}^0 \cup \mathcal{S}^0$ & - &15.2 & 11.9 & - &  54.5 & 55.4   \\
    \cline{3-9}
    & & $\mathcal{S}^X \cup \mathcal{A}^0 \cup \mathcal{S}^0$ & - &14.0 & 18.8 & - & 50.6 & 60.1 \\
    \bottomrule
    \end{tabular}
    }
\vspace*{-1\baselineskip}
\end{table*}

\noindent\textbf{Baselines.}
We mainly focus on comparing against the state-of-the-art FSOD baselines: Faster-RCNN~\cite{ren2015faster}, SSD~\cite{liu2016ssd}, and Sparse DETR~\cite{roh2021sparse}; WSOD baselines: WSDDN~\cite{bilen2016weakly}, OICR~\cite{tang2017multiple}, C-MIDN~\cite{gao2019c}, WSOD2~\cite{zeng2019wsod2}, MIST(+Reg)~\cite{ren2020instance}, CASD~\cite{huang2020comprehensive}, W2N~\cite{huang2022w2n}, and D2DF2WOD~\cite{wangD2DF2WOD}; SSOD baselines: BCNet~\cite{pan2019low}, OAM~\cite{biffi2020many}, Unbiased teacher v2~\cite{liu2022unbiased}, and Label Matching~\cite{chen2022label}; and ALOD baselines: BAOD~\cite{pardo2021baod}, SSDGMM~\cite{choi2021active}, NALAE~\cite{elezi2022not}, Active Teacher~\cite{mi2022active}, and BiB~\cite{vo2022active}. 

\subsection{Main Results}\label{sec:main_result}
In total, we annotate $5\%$ of training images on VOC2007 and RealPizza10, $1\%$ of training images on COCO2014. We perform five annotation cycles
with a budget of $B=50$ images per cycle on VOC2007 and RealPizza10, and $B=160$ images per cycle on COCO2014. 

We compare our method \alwod to state-of-the-art FSOD ($n/N = 100\%$ FA data in $\mathcal{S}$), WSOD  ($n/N= 0\%$ FA data in $\mathcal{S}$), and SSOD methods ($n/N = 10\%$ or $n/N = 5\% / 1\% / 5\%$ FA data in $\mathcal{S}$), where $n$ is the total number of fully-labeled samples.
\cref{tab:voc_coco_baseline} summarizes the detection results on different benchmarks.

\noindent\textbf{Our method outperforms the ALOD baselines.} As shown in~\cref{tab:voc_coco_baseline}, on VOC2007 our method reaches 68.0\% AP50, outperforming SSDGMM~\cite{choi2021active} and BiB~\cite{vo2022active} by 39.2\% and 7.4\% absolute points, using the same number of fully-annotated images. On the more challenging COCO2014, our method reaches 38.4\% AP50, outperforming BiB~\cite{vo2022active} by 5.2\%. Using the same number of annotated training images, our detection performance benefits from the combination of semi-supervised learning and active learning compared with SSDGMM and WSOD-based BiB. On RealPizza10, our method outperforms SSDGMM~\cite{choi2021active} using 80\% fully-annotated data by 14.5\% using only 5\% fully-annotated data. The auxiliary set is comparable with the large fully-labeled real data. \textbf{Our method also outperforms the SSOD baselines.} On VOC2007 in~\cref{tab:voc_coco_baseline}, we observe that our method improves by over $n/N=10\%$ SSOD baselines, while using only $n/N=5\%$ full-annotated data. \textbf{Our method also outperforms the WSOD baselines.} Compared to WSOD baselines, our method obtains significantly better results, across the three benchmarks, with only a small amount of full-annotated data. \textbf{Our method generalizes across different datasets,} performing particularly well on datasets that contain multiple and varied object instances per image, \ie, RealPizza10 and COCO2014.
\textbf{Our method generalizes to different backbones.} Our method approaches to FSOD state-of-the-art Sparse DETR~\cite{roh2021sparse} by 3.4\% AP50 difference with only 5\% fully-annotated training images on RealPizza10 using ResNet50 backbone. 
\textbf{Our method generalizes to different detectors.} We compare Faster-RCNN with Sparse DETR for both $\mathcal{G}^{stu}$ and $\mathcal{G}^{tea}$. They both perform similarly. Hence, the performance gain is largely not affected by the detector architecture. 
\alwod shows a better trade-off between detection performance and annotation effort than FSOD, WSOD, and SSOD. Qualitative results can be found in \cref{sec:qualitative}.

\subsection{Ablation Study}\label{sec:ablation}
\noindent\textbf{Impact of $\mathcal{S}^X$.} We analyze the impact of our warm-start active learning  approach using the auxiliary set $\mathcal{S}^X$, described in~\cref{sec:init_learn}, on RealPizza10 and VOC2007 datasets. \alwod first annotates randomly selected images $\mathcal{A}^0$, then creates $\mathcal{S}^X$ constructed from task-specific augmentations of these fully-labeled images $\mathcal{A}^0$. The initial student-teacher model is trained on weakly-annotated $\mathcal{S}^0$ and fully-annotated $\mathcal{A}^0$ or $\mathcal{S}^X$. In~\cref{tab:auxiliary_ab} $\mathcal{S}^X \cup \mathcal{A}^0 \cup \mathcal{S}^0$ denotes that a model is trained on the fully-annotated auxiliary images $\mathcal{S}^X$ coupled with weakly-annotated real images $\mathcal{S}^0$ in a semi-supervised manner, where $\mathcal{A}^0$ is not directly used for training, while used for creating $\mathcal{S}^X$. To our knowledge, this is the first use of this concept in active learning. We compare our framework with SSDGMM~\cite{choi2021active}, and BiB~\cite{vo2022active} using different training sets. In our framework, we only use fully-labeled images $\mathcal{A}^0$ or auxiliary domain $\mathcal{S}^X$ to initialize the student and teacher networks. In SSDGMM~\cite{choi2021active}, the initial FSOD model is only trained on fully-annotated images $\mathcal{A}^0$ or auxiliary domain $\mathcal{S}^X$. In BiB, we first pre-train the base weakly-supervised detector MIST~\cite{vo2022active} on $\mathcal{S}^0$. Then we fine-tune MIST either on $\mathcal{A}^0$ selected by BiB AL strategy or $\mathcal{S}^X$ constructed from $\mathcal{A}^0$.

\cref{tab:auxiliary_ab} shows the key role of $\mathcal{S}^X$ in lifting the initial detection performance of \alwod from 11.9\% to 18.8\% in terms of AP50 on RealPizza10. As shown in~\cref{tab:auxiliary_ab} our approach significantly outperforms the initialization strategy of SSDGMM~\cite{choi2021active} by utilizing the auxiliary domain.

We also analyze the impact of our auxiliary domain $\mathcal{S}^X$ in the BiB framework. As shown in~\cref{tab:auxiliary_ab}, in \alwod $\mathcal{S}^X$ helps to initialize the student-teacher model. $\mathcal{S}^X$ improves the performance of MIST, while due to the domain gap between $\mathcal{S}^X$ and $\mathcal{S}^0$, the improvement is less than the improvement using $\mathcal{A}^0$.  

\begin{figure}
        \centering
        \begin{subfigure}{\linewidth}
            \centering
            \includegraphics[trim={0cm 0cm 0.6cm 1cm},clip, scale=0.6]{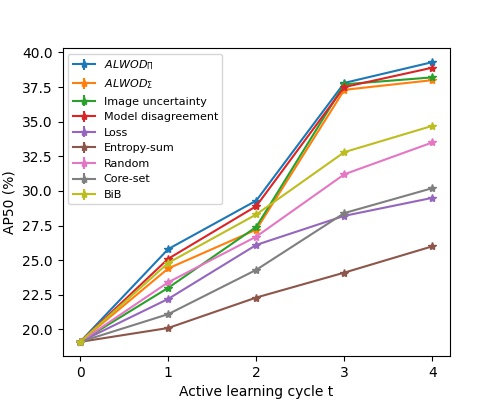}
            \caption{RealPizza10}            \label{fig:al_strategies_pizza}  
            \end{subfigure}
        \vfill
        \begin{subfigure}{\linewidth}
            \centering
            \includegraphics[trim={0.4cm 0cm 0.6cm 1cm}, clip, scale=0.6]{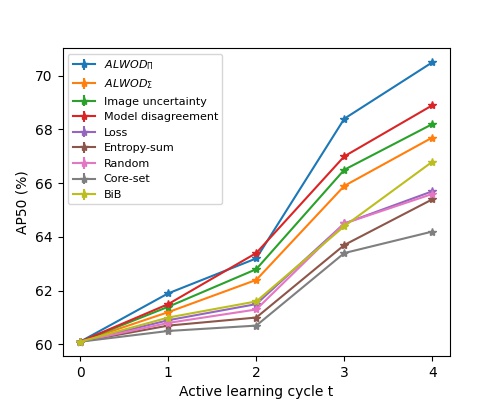}
            \caption{VOC2007}
        \end{subfigure}
        \vfill
        \begin{subfigure}{\linewidth}
            \centering
            \includegraphics[trim={0.4cm 0cm 0.6cm 1cm}, clip, scale=0.6]{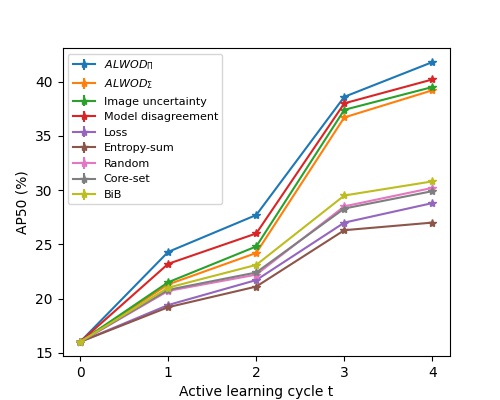}
        \label{fig:al_strategies_coco}
            \caption{COCO2014}
      \end{subfigure}
      \vspace*{-1.5\baselineskip}
        \caption{Detection performance across different active learning strategies in our framework on the three benchmarks using ResNet50 backbone.}
        \label{fig:al_strategies}
        \vspace*{-2.0\baselineskip}
    \end{figure}

\noindent\textbf{Comparison of active learning strategies.}
We investigate the performance of our semi-supervised framework on three benchmarks under different acquisition functions using our proposed annotation strategy. We complete $T=5$ annotation cycles with
a budget of $B=50$ images per cycle on VOC2007 and RealPizza10, $B=160$ images per cycle on COCO2014 datasets. We report the performance using the average
of AP50 over three repetitions. As shown in~\cref{fig:al_strategies}, across the three benchmarks, our acquisition method $\alwod_{\Pi}$ consistently achieves improvement over other strategies. The performance of BiB trails our $\alwod_{\Pi}$, since BiB strategy aims to select the ``best'' training samples to ``fix'' the mistakes of the base weakly-supervised detector~\cite{vo2022active}, while \alwod disagreement-based AL strategy leverages the key property of the student-teacher networks. Entropy-sum obtains significantly worse results than other strategies on RealPizz10 and COCO2014 datasets as shown in~\cref{fig:al_strategies_pizza,fig:al_strategies_coco}.
Core-set~\cite{sener2017active} and loss~\cite{yoo2019learning} underperform uniform sampling. The performance of uniform sampling is always worse than our proposed functions in each AL cycle. At $t=3$, we see a significant improvement compared with the previous steps. As shown in~\cref{fig:al_strategies}, our product strategy $\alwod_{\Pi}$ for the final fused acquisition function exceeds the sum strategy $\alwod_{\Sigma}$. Per class results can be found in \cref{sec:add_results}.

\noindent{\bf Comparison of annotation strategies.} As shown in ~\cref{tab:alstrat}, our final fused acquisition function $\alpha_{\Pi}$ combined with our ``selecting the box'' annotation strategy obtains comparable results to the traditional ``drawing the box'' annotation strategy by only a 0.6\% difference in terms of AP50.
\begin{table}[hbt!]
\caption{Ablation study on affect annotation tools based on our semi-supervised framework on RealPizza10 dataset using ResNet50 backbone.}
\label{tab:alstrat}
\footnotesize
\resizebox{\columnwidth}{!}{
\begin{tabular}{cccccc}
\toprule
Labeling & \multicolumn{5}{c}{AP50 (\%)} \\
Cycle (t) & 0 & 1 & 2 & 3 & 4 \\
\midrule
Drawing & 19.1 & 27.1 &37.3 & 38.8 & 39.9\\
Selecting (\alwod) & 19.1 & 25.8 & 29.3 & 37.7 & 39.3\\
\bottomrule
\end{tabular}
}
\vspace*{-1.5\baselineskip}
\end{table}

%% file: 5_conclusion.tex
\section{Discussion and Conclusion}
We propose a new approach to boost the detection performance of weakly-supervised object detectors by combining semi-supervised learning and active learning. We introduce a simple yet effective image generator to create large auxiliary fully-annotated data by leveraging few fully-annotated real images to warm-start active learning. Our framework introduces a novel acquisition function based on the fusion of the student-teacher OD model disagreement and the traditional image uncertainty, combined with an effective, low-effort annotation strategy. Empirical evaluations show that our method significantly outperforms the state-of-the-art on several key benchmarks and is particularly adept at tackling challenging multi-object, multi-class scenarios such as those in COCO2014 and RealPizza10 datasets.
\noindent\textbf{Limitation.} Our annotation strategy requires the framework to automatically select certain object proposals on each image for manual checking. The default selection criteria may either introduce noisy or false positive bounding boxes or ignore bounding boxes with true objects in them; this may negatively affect the annotation quality or the annotation speed and, subsequently, the OD accuracy.

\noindent{\bf Acknowledgement:} This material is based upon work supported by 
NSF IIS Grant \#1955404.

%% file: 6_appendix.tex
This appendix is organized as follows. In~\cref{sec:PL}, we provide the implementation details of pseudo-labeling filtering included in the framework of \alwod. Afterward, we provide additional implementation details of the auxiliary image generator in~\cref{sec:auxilary_domain}, and additional implementation details of the annotation tool in~\cref{sec:annotation_sup}. In~\cref{sec:add_results}, we provide additional experimental results. In~\cref{sec:qualitative}, we provide qualitative evaluation results.

\section{Pseudo-Labeling Filtering}
\label{sec:PL}
Our semi-supervised object detector is composed of a student detection network $\mathcal{G}^{stu}(\mathbf{X}|\theta_{stu})$ and a teacher detection network  $\mathcal{G}^{tea}(\mathbf{X}|\theta_{tea})$. Both $\mathcal{G}^{stu}$ and $\mathcal{G}^{tea}$ are transformer-based object detectors~\cite{carion2020end,roh2021sparse}. We adopt Sparse DETR~\cite{roh2021sparse} for both student and teacher networks since Sparse DETR enhanced the efficiency of DETR~\cite{carion2020end} and improved the detection performance on small objects datasets. We apply both weak $\mathrm{U}_{weak}(\cdot)$ and strong $\mathrm{U}_{strong}(\cdot)$ augmentation to the data. 
At the initial stage ($t = 0 $), we train the student network using $\mathrm{U}_{weak}(\mathcal{S}^X) \cup \mathrm{U}_{strong}(\mathcal{S}^X)$ as well as the pseudo-labeled $\mathrm{U}_{strong}(S^0)$, with labels proposed by $\mathcal{G}^{tea}(\mathrm{U}_{weak}(S^0))$\ and filtered by $\phi(\cdot)$:
\begin{equation}
    \theta_{stu}^0 \leftarrow \min \sum_{(\mathbf{X},\mathbf{Y}) \in \mathcal{T}^0 } \mathcal{L}(\mathcal{G}^{stu}(\mathbf{X}|\theta_{stu}),\mathbf{Y}),
    \label{eq:learnstu0}
\end{equation}
where
\begin{equation}
    \mathcal{T}^0 = \mathrm{U}_{weak}(\mathcal{S}^X) \cup \mathrm{U}_{strong}(\mathcal{S}^X) \cup \phi(\mathcal{G}^{tea}(\mathrm{U}_{weak}(S^0))),
    \label{eq:learnstu_train0}
\end{equation}
and $\mathcal{L}$ is the classification and bounding box regression loss used in transformer-based detector~\cite{carion2020end,roh2021sparse}. 
At the active learning cycle $(t >0)$, we train the student network using $\mathrm{U}_{weak}(\mathcal{S}^t) \cup \mathrm{U}_{strong}(\mathcal{S}^t)$ as well as the pseudo-labeled $\mathrm{U}_{strong}(S^t)$, with labels proposed by $\mathcal{G}^{tea}(\mathrm{U}_{weak}(S^t))$\ and filtered by $\phi(\cdot)$:
\begin{equation}
    \theta_{stu}^t \leftarrow \min \sum_{(\mathbf{X},\mathbf{Y}) \in \mathcal{T}^t } \mathcal{L}(\mathcal{G}^{stu}(\mathbf{X}|\theta_{stu}),\mathbf{Y}),
    \label{eq:learnstut}
\end{equation}
where
\begin{equation}
    \mathcal{T}^t = \mathrm{U}_{weak}(\mathcal{S}^t) \cup \mathrm{U}_{strong}(\mathcal{S}^t) \cup \phi(\mathcal{G}^{tea}(\mathrm{U}_{weak}(S^t))).
    \label{eq:learnstu_traint}
\end{equation}

Our pseudo-labeling filter $\phi(\cdot)$ supports two annotation forms: 1) {\em weakly annotated} $\mathbf{Y}^w_{j}$, where $j\in W^t$ and 2) {\em fully annotated} $\mathbf{Y}^f_{j}$, where  $j\in F^t$. The filter $\phi(\cdot)$ is applied to predicted bounding boxes $\mathbf{\hat  Y}$ and ground-truth labels $\mathrm{U}_{weak}(\mathbf{Y})$ associated with image $\mathrm{U}_{weak}(\mathbf{X})$ , where $\mathbf{\hat  Y}$ = $ \mathcal{G}^{tea}(\mathrm{U}_{weak}(\mathbf{X})) = \{\mathbf{\hat y}_k\}_{k=1}^{K}$, and K is the number of object queries. We formulate the pseudo-label filtering problem as a bipartite matching problem between  $\mathbf{\hat  Y}$ and  $\mathrm{U}_{weak}(\mathbf{Y})$~\cite{wang2022omni} using a matching function across a permutation of $K$ elements with the lowest cost as following:
\begin{equation}
    \hat \sigma = \argmin_{\sigma \in \wp_{K}} \sum_{k=1}^{K}\mathcal{L}_{match}(\mathbf{\hat  y}_{\sigma(k)},\mathrm{U}_{weak}(\mathbf{y}_k)),
\label{eq:detr_los}
\end{equation}
where $\mathcal{L}_{match}(\mathbf{\hat  y}_{\sigma(k)},\mathrm{U}_{weak}(\mathbf{y}_k))$ is an annotation pair-wise matching score between ground-truth label $\mathrm{U}_{weak}(\mathbf{y}_k)$ and teacher prediction $\mathbf{\hat  y}$ with index $\sigma(k)$.
It is computed efficiently with the Hungarian algorithm~\cite{carion2020end,kuhn1955hungarian}. Specifically, for different types of annotations $\mathrm{U}_{weak}(\mathbf{y}_k)$, we define different $\mathcal{L}_{match}$ loss functions. 

\noindent\textbf{The ground-truth label is weakly annotated.} The ground-truth label contains only the classes of objects present in that image but not the objects' locations, \ie , $\mathbf{Y}^w_{j} = \{\mathbf{y}_{k}\}_{k=1}^{n^w} =  \{c_{k}\}_{k=1}^{n^w}$, where $1 \leq n^w \leq C$ is the number of object classes in that image. To address the matching problem, we first predict the count $n_k$ of the class $c_k$,
\begin{equation}
    n_k = \max (1, |\{o|pr_{o}^{c_k} > \delta, o \in [1, K]\}|),
\end{equation}
where $pr_{o}^{c_k}$ is the probability of assigning the $o$-th prediction to class $c_k$. The predicted count $n_k$ is the number of predictions that pass the confidence threshold $\delta$. In our work, $\delta$ is 0.7. Since there is at least one object per ground-truth class, the minimal value of $n_k$ is 1. Each class $c_k$ will be repeated $n_k$ times, and the total number of ground-truth labels will be $\sum_{k=1}^{n^{w}} n_k$. Then we find the best matched bounding boxes $\hat {\mathbf{b}}_{\hat \sigma(k)}$ by
$\mathcal{L}_{match}(\mathbf{\hat  y}_{\sigma(k)},\mathrm{U}_{weak}(\mathbf{y}_k))$ in~\cref{eq:detr_los}, where: 
\begin{equation}
    \mathcal{L}_{match}(\mathbf{\hat  y}_{\sigma(k)},\mathrm{U}_{weak}(\mathbf{y}_k)) = 1 - pr_{\sigma(k)}^{c_k},
\end{equation}
where $\sigma (k) \in \{1, \ldots, K\}$ and $k \in \{1, \ldots, \sum_{k=1}^{n^{w}} n_k\}$.
After bipartite matching, we generate the pseudo-label $\widetilde{\mathbf{y}_{k}} = \{( \hat {\mathbf{b}}_{\hat \sigma (k)},c_{k}, p_{k})\}_{k=1}^{\sum_{k=1}^{n^{w}} n_k}$, where $\hat \sigma (k) \in \{1, \ldots, K\}$ is the index of matched predicted box to the $k$-the ground-truth label, $\hat {\mathbf{b}}_{\hat \sigma(k)}$ is the predicted box, $c_k$ is the available ground-truth class, and $p_k = pr_{\hat \sigma(k)}^{c_k}$ is the pseudo label quality score.

\noindent\textbf{The ground-truth label is fully annotated.} The ground-truth label $\mathbf{Y}^f_{j} = \{\mathbf{y}_{k}\}_{k=1}^{n^f} =  \{(\mathbf{b}_{k},c_{k},p_{k})\}_{k=1}^{n^f}$ contains localization bounding box $\mathbf{b}_{k}\in{\mathbb{R}}^{4}$, class-label $c_{k} \in \{1,\ldots,C \}$, and the bounding box quality score $p_{k} \in \{ 1, 0\}$, for each of the $n^f$ objects labeled in that image. The score $p_{k}$ corresponds to a subjective (annotator) notion of whether the bounding box $\mathbf{b}_{k}$ is precise, $p_k = 1$, where the bounding box $\mathbf{b}_{k}$ largely overlaps with the true object ($\mathrm{IoU}\geq 0.9)$, or imprecise, $p_k = 0$, where $0.5<\mathrm{IoU}<0.9$. For a precise bounding box, where $p_k = 1$, the pseudo label $\widetilde{\mathbf{y}_{k}}$ is exactly the same as $\mathbf{y}_{k}$. For an imprecise bounding box $\mathbf{b}_k$, where $p_k=0$, we first find the best matched predicted
bounding boxes $\hat {\mathbf{b}}_{\hat \sigma(k)}$ by $\mathcal{L}_{match}(\mathbf{\hat  y}_{\sigma(k)},\mathrm{U}_{weak}(\mathbf{y}_k))$ in~\cref{eq:detr_los}, where :
\begin{equation}
\begin{split}
    \mathcal{L}_{match}(\mathbf{\hat  y}_{\sigma(k)},\mathrm{U}_{weak}(\mathbf{y}_k)) = \lambda_{iou} \mathcal{L}_{iou} (\mathbf{\hat b}_{\sigma (k)},\mathbf{b}_{k}) \\
    + \lambda_{L_{1}} ||\mathbf{\hat b}_{\sigma (k)} - \mathbf{b}_{k}||_1,
\end{split}
\end{equation}
$\mathcal{L}_{iou}$ is the generalized IoU loss~\cite{carion2020end}, $\lambda_{iou}$ and $\lambda_{L_{1}}$ are trade-off parameters. In our work, $\lambda_{iou}$ is 2 and $\lambda_{L_{1}}$ is 5. Then we generate the final pseudo-label $\widetilde{\mathbf{y}_{k}} = \{(\mathbf{b}^*_{\hat \sigma(k)},c_{k}, p_{k})\}_{k=1}^{n^{f}}$, where $\mathbf{b}^{*}_{\hat \sigma(k)}$ is generated by interpolating the coordinates of
the imprecise $\mathbf{b}_k$ and the matched boxes $\hat {\mathbf{b}}_{\hat \sigma(k)}$.

\section{Auxiliary Image Generator}
\label{sec:auxilary_domain}
\begin{figure*}[ht!]
\centering
\includegraphics[trim={0cm 0cm 0cm 0},clip, scale=0.5]{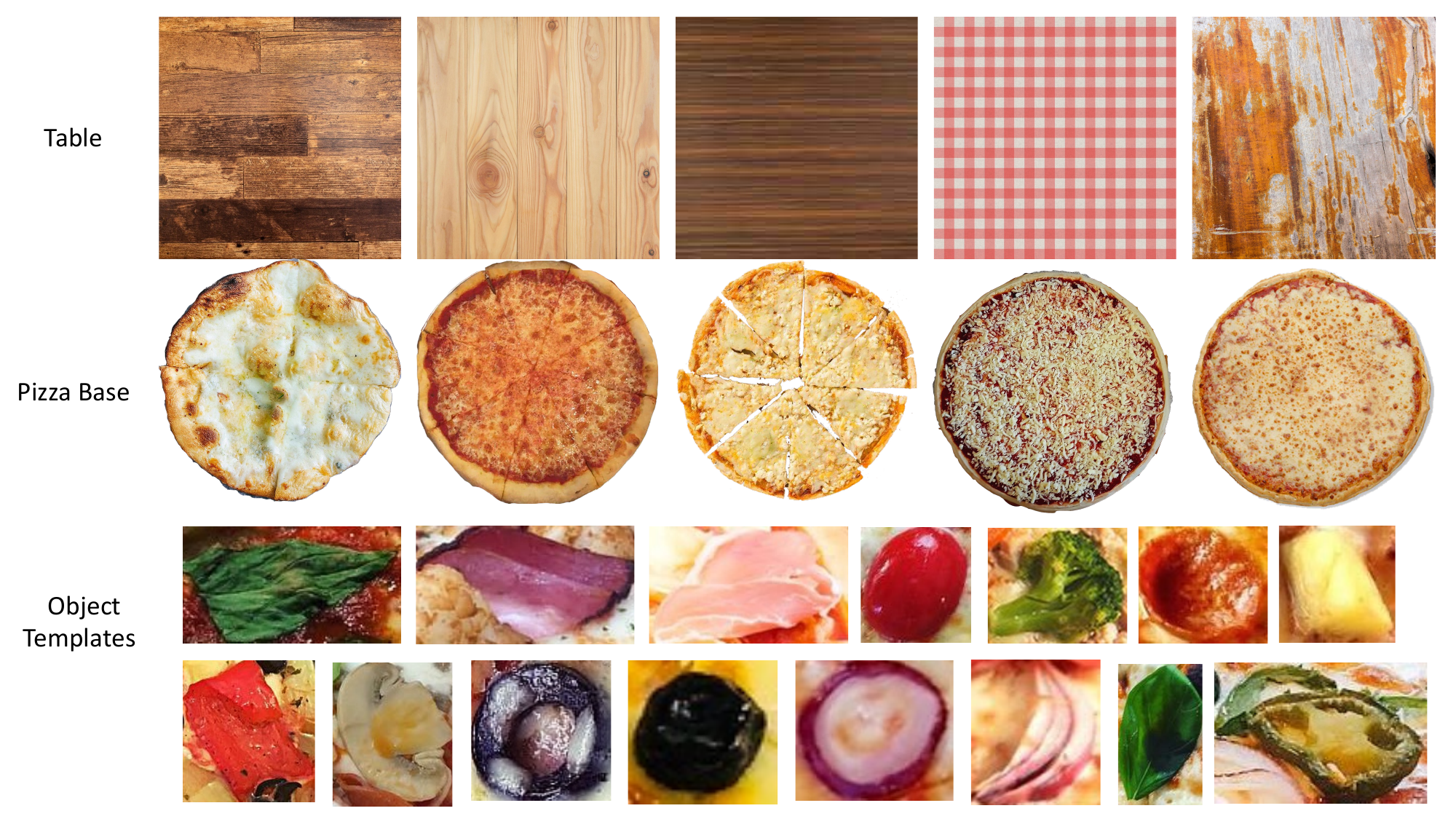}
\caption{Examples of table images, pizza base images, and object templates used for creating auxiliary images on RealPizza10 dataset.
}
\label{fig:pizza_image}
\end{figure*}

\begin{figure*}[ht!]
\centering
\includegraphics[trim={0cm 2.2cm 0cm 0},clip, scale=0.53]{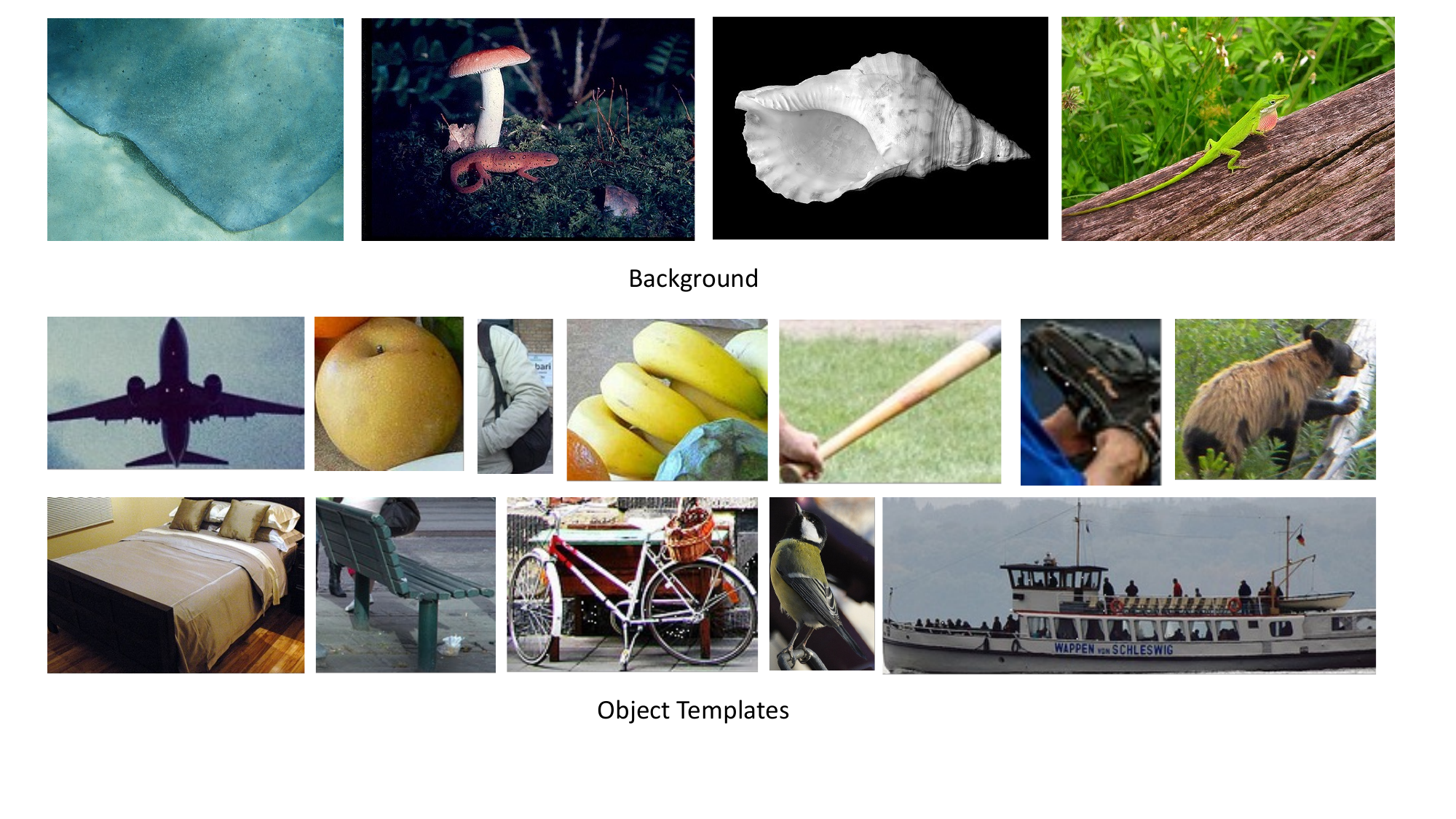}
\caption{Examples of background images and object templates used for creating auxiliary images on COCO2014 dataset.
}
\label{fig:coco_image}
\vspace{2cm}
\end{figure*}

The image generator creates synthetic images by composing background images
and object templates. For RealPizza10~\cite{wangD2DF2WOD} dataset, the background images include pizza base images and table images as shown in ~\cref{fig:pizza_image}. We download ten table images and ten pizza base images from Google, using several pizza-related hashtags. For VOC2007~\cite{everingham2010pascal} and COCO2014~\cite{lin2014microsoft} datasets, the background images are nature images as shown in~\cref{fig:coco_image}. We randomly select 300 images from the ImageNet~\cite{russakovsky2015imagenet} dataset, which do not contain any classes in COCO2014 dataset. 150 of these images are used as background images for VOC2007 dataset, and the remaining images are used as background images for COCO2014 dataset. The object templates are created by cropping the object instances of fully-annotated images in $A^0$, which are randomly selected from $S^0$.

\section{Annotation Procedure and Tool}\label{sec:annotation_sup}
In most object detection labeling software, the user's task is to, for an \textit{un-annotated image} proposed by the system (e.g., every image in a dataset, randomly selected image, or selected by an AL algorithm), draw tight bounding boxes around the objects to be detected, and select categories for each bounding box. To speed up the process of labeling and to reduce the human effort, we develop a new annotation tool. 

Each image contains a large number of predicted bounding boxes with predicted class labels generated from both the student and the teacher networks in \alwod. We first group all predictions into different clusters, such that the predictions in the same cluster are overlapped the most.
In each cluster, the cluster center is the largest predicted bounding box and the remaining predictions are fully covered by the cluster center. Subsequently, we retain the cluster center predictions and discard the remaining predictions to prevent the detection network from concentrating too much on parts of objects instead of whole objects.
Secondly, we adopt non-maximum
suppression (NMS) on the cluster center predictions based on their predicted classes. We fix the IoU threshold for NMS at 0.75. Thirdly, we remove all the predicted bounding boxes that have low confidence scores. The confidence score threshold is 0.3. The remaining predictions are saved as proposals which are given to the annotators. The proposals generated from the teacher network are denoted by ``D3'', and the proposals generated from the student network are denoted by ``D4'' in our annotation tool.

\begin{figure*}[ht!]
\centering
 \includegraphics[width=0.9\textwidth]{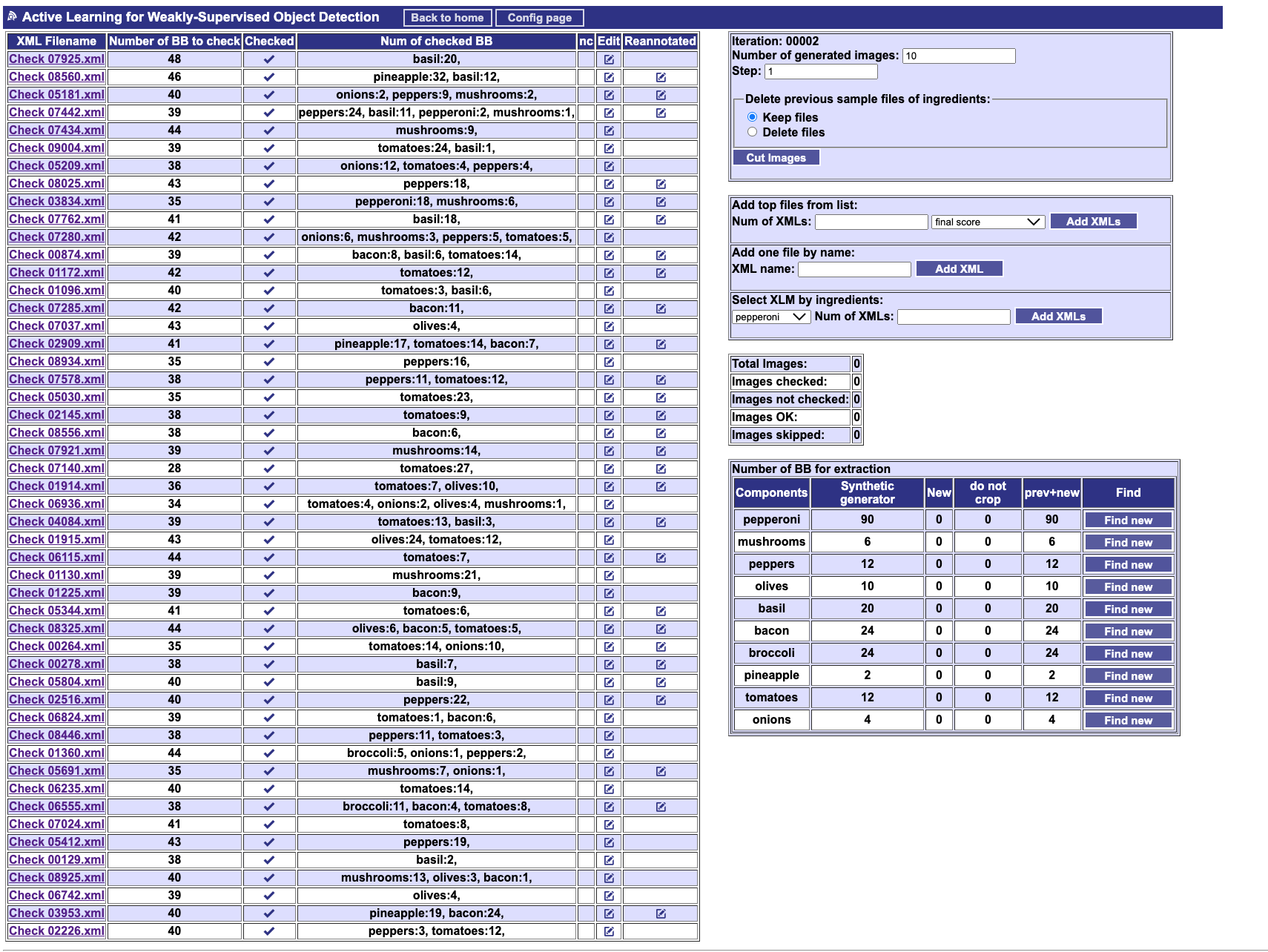} 
 \caption{The annotation tool presents a list of images that need to be examined by an annotator, sorted according to the final fused acquisition score.
 }
\label{xml_list}
\end{figure*}

\begin{figure*}[th!]
\centering
 \includegraphics[width=0.9\textwidth]{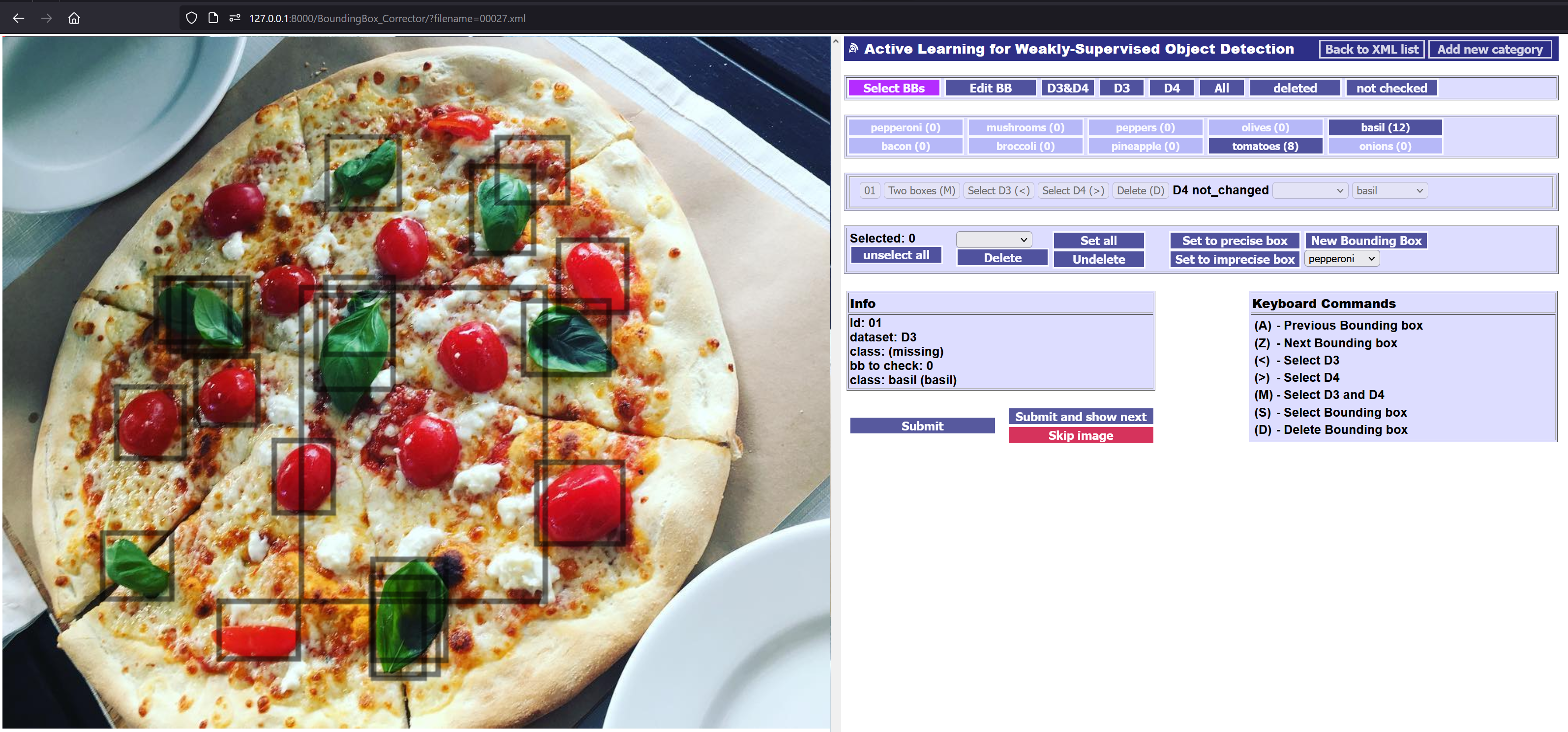} 
 \caption{Image with predictions generated by $\mathcal{G}_{tea}(\cdot | \theta_{tea}^{k-1})$ and $\mathcal{G}_{stu}(\cdot | \theta_{stu}^{k-1})$ networks, filtered by clustering, NMS and the confidence score threshold. The predictions generated from the teacher network are denoted by ``D3'', and the predictions generated from the student network are denoted by ``D4'' in our annotation tool.
 }
\label{web_tool_before}
\end{figure*}

\begin{figure*}[th!]
\centering
 \includegraphics[width=0.9\textwidth]{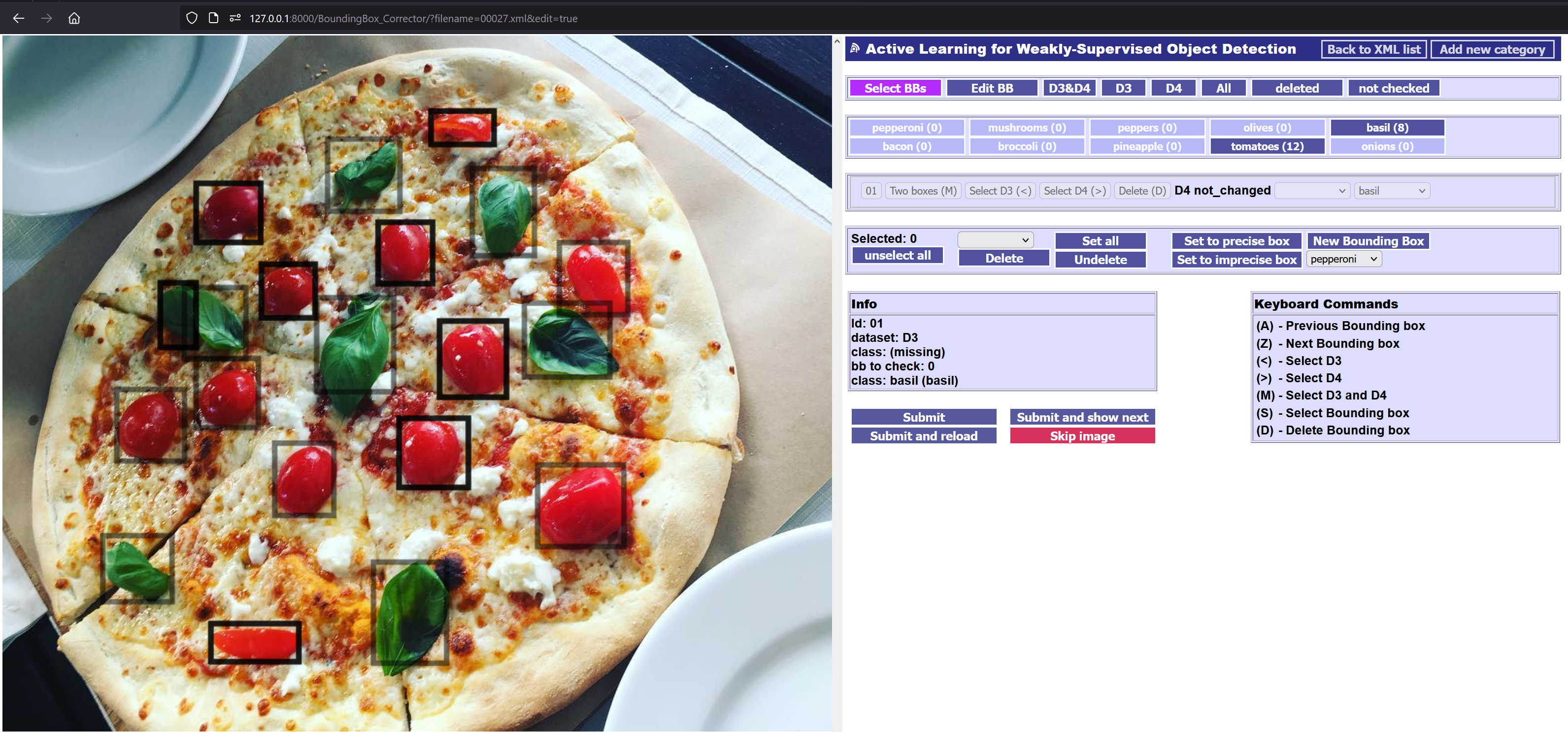} 
 \caption{Image with annotations after (1) manually correcting the object classes, (2) removing unselected bounding boxes, and (3) adding new object bounding boxes. Newly added bounding boxes are marked in black.
 }
\label{web_tool_after}
\end{figure*}

All the selected images with proposals in $A^t$ are first loaded into our annotation tool as shown in~\cref{xml_list}. At the beginning, as shown in \cref{web_tool_before}, each image includes multiple proposed bounding boxes with class labels. As shown in~\cref{web_tool_after}, the annotators are asked to: (1) select the bounding boxes from proposals that overlap with true objects ($>50\%$ IoU) and include at
least one of the four extreme points (top, bottom, left-most, right-most), (2) correct the bounding box categories, and (3) assess the bounding box quality: {\em precise} or {\em imprecise} bounding boxes. The remaining unselected
bounding boxes are removed. If there is no bounding box over an actual object, the annotators directly draw a tight bounding box and select the object label.

\begin{figure*}[th!]
\centering
 \includegraphics[width=0.9\textwidth]{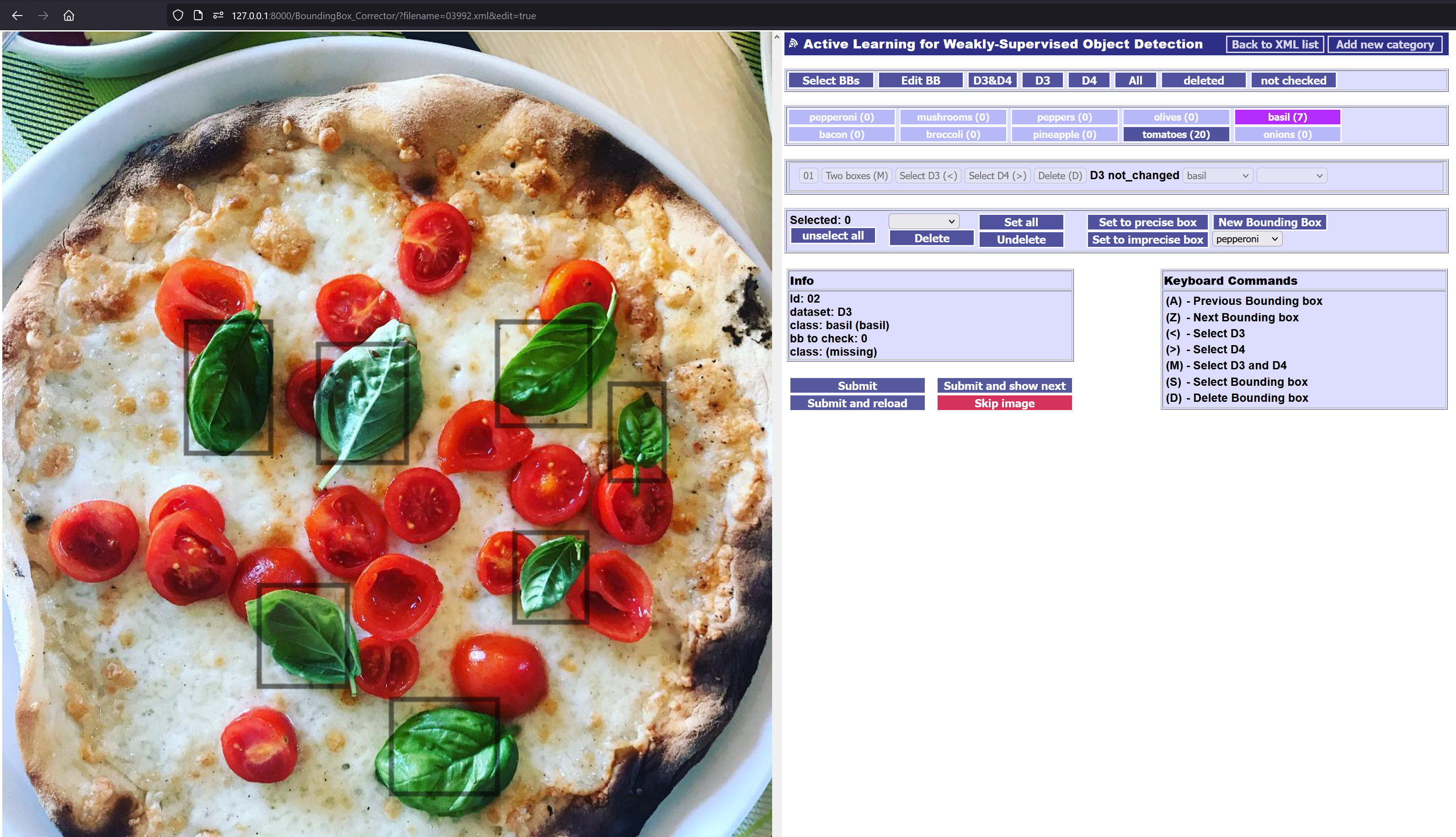} 
\caption{The annotator checks all the bounding boxes predicted as basil, when they click the basil button at the right side of interface (purple).
 }
\label{web_tool_basil}
\end{figure*}

The web tool allows the annotator to display certain classes of bounding boxes as shown in~\cref{web_tool_basil}. In this way, it is easy to notice when one of the bounding boxes is assigned to the wrong category. The tool allows the user to select multiple bounding boxes at the same time, and simultaneously change their categories or delete them, which significantly speeds up corrections. Our annotation tool also allows the annotator to add new classes, when the user observes objects outside of the existing data categories.

\section{Additional Results}
\label{sec:add_results}

\begin{figure*}[hbt!]
\centering
 \includegraphics[trim={0cm 6cm 9cm 3cm},clip, scale=0.7]{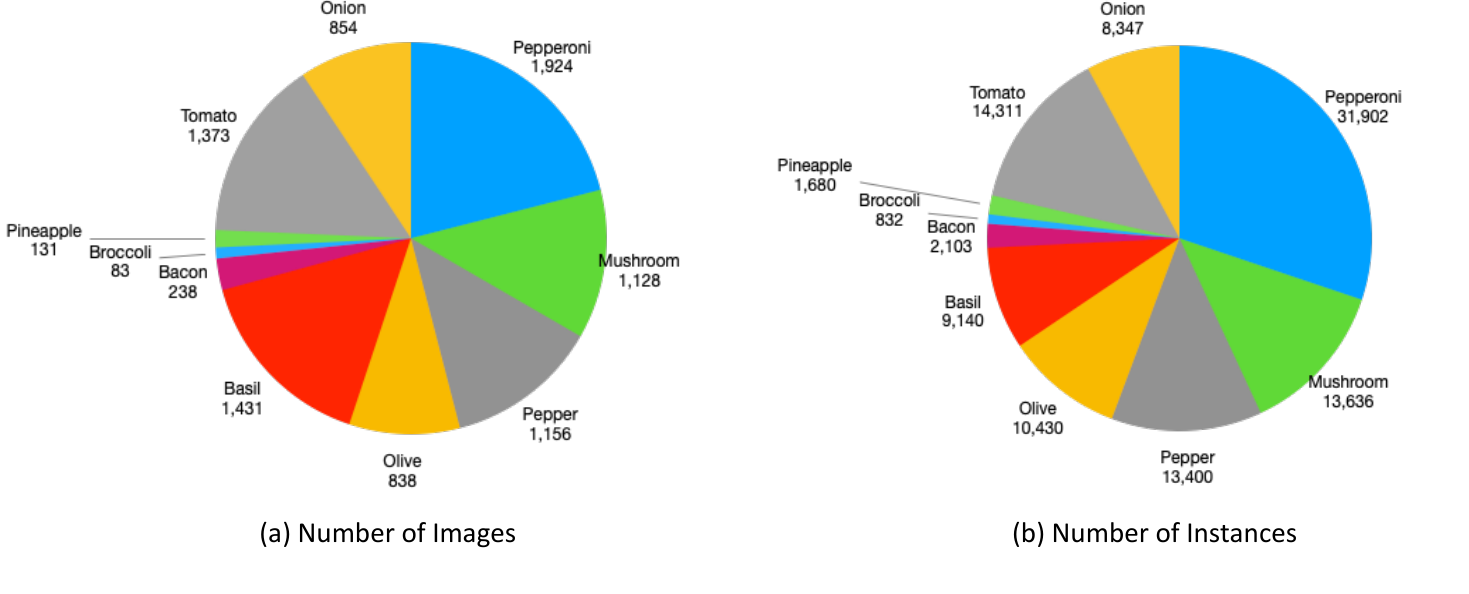}
 \caption{Statistics of RealPizza10 dataset: (a) the number of images, (b) the number of instances.
 }
\label{fig:num_instance_pizza}
\end{figure*}

\begin{table*}[hbt!]
    \centering
    \caption{Results (mean 
    AP50 over all classes and per class AP50 in \%) for different methods on RealPizza10 dataset. $n$ is the total number of fully-labeled samples and $N$ is the total number of samples in RealPizza10 dataset.
    Red figures denote the best performing {\em non-FSOD method} and blue figures denote the second best performing {\em non-FSOD method} using the same backbone. Our method significantly boosts detection performance in most classes. SSDGMM considers 80\% or 5\% of fully-annotated RealPizza10 data, and \alwod considers 5\% fully-annotated RealPizza10 data.
    }
    \vspace{0.3cm}
    \label{tab:pizza_main}
    \footnotesize
    \resizebox{2.0\columnwidth}{!}{
    \begin{tabular}{c@{\hspace{5mm}}|c@{\hspace{5mm}}|c@{\hspace{5mm}}|c@{\hspace{2mm}}|c@{\hspace{2mm}}|c@{\hspace{2mm}} c@{\hspace{2mm}} c@{\hspace{2mm}} c@{\hspace{2mm}} c@{\hspace{2mm}} c@{\hspace{2mm}} c@{\hspace{2mm}} c@{\hspace{2mm}} c@{\hspace{2mm}} c@{\hspace{2mm}} c@{\hspace{2mm}}}
    \toprule
    \multirow{2}{*}{Backbone} & \multirow{2}{*}{Setting} & \multirow{2}{*}{Method}& \multirow{2}{*}{$n/N$} & Mean & \multicolumn{10}{c}{AP50} \\
    & & & & AP50 & Pepperoni & Mushroom & Pepper & Olive & Basil & Bacon & Broccoli & Pineapple & Tomato & Onion\\    
    \midrule
    VGG16 & \multirow{5}{*}{FSOD} &
    Faster-RCNN~\cite{ren2015faster}  & 100\% & 39.1 & 74.2 & 31.6 & 31.4 & 57.8 & 58.4 & 11.4 & 34.2 & 11.4 & 51.6 & 29.0 \\
    VGG16 & & Sparse DETR~\cite{roh2021sparse}  & 100\% & 41.2 & 79.4 & 39.6 & 32.6 & 56.8 & 62.5 & 16.8 & 38.9 & 7.6 & 53.4 & 24.8 \\
    ResNet50 & &
    Faster-RCNN~\cite{ren2015faster}  & 100\% & 40.2 & 73.9 & 37.7 & 29.3 & 56.3 & 56.4 & 14.9 & 43.8 & 11.4 & 50.0 & 28.1 \\
    ResNet50 & & Sparse DETR~\cite{roh2021sparse} & 100\% & 42.7 & 81.3 & 41.6 & 30.1 & 58.4 & 64.1 & 12.7 & 41.4 & 14.3 & 56.3 & 26.5 \\
    Swin-T & 
    & Sparse DETR~\cite{roh2021sparse} & 100\% & 43.8 & 80.5 & 44.2 & 32.1 & 60.1 & 64.7 & 14.4 & 41.0 & 16.6 & 54.4 & 29.8 \\
    \midrule
    \midrule
    \multirow{6}{*}{VGG16} & WSOD
    & OICR~\cite{tang2017multiple} & 0\% &  4.7
    & 0.2 & 1.3 & 4.5 & 0.1 & 0 & 8.8 & \color{blue}19.4 & \color{blue}{11.0} & 1.0 & 0.8 \\ 
    & WSOD & CASD~\cite{huang2020comprehensive} & 0\% & 12.9 & 12.7 & 19.5 & 14.8 & 10.5 & 13.7 & \color{blue}{10.4} & 10.1 & \color{red}14.5 & 11.7 & 10.7 \\
    & WSOD & ${\text{D2DF2WOD}}$\cite{wangD2DF2WOD} & 0\% & \color{blue}{25.1} & 43.9 & \color{blue}{35.1} & \color{blue}{15.0} & 27.3 & 41.8 & 9.2 & 12.5 & 8.5 & 28.4 & \color{red}29.2 \\
    & ALOD & SSDGMM ~\cite{choi2021active} & 80\% & 23.4 & \color{blue}62.5 & 20.7 & 14.5 & \color{blue}{32.7} & \color{blue}{45.2} & 4.9 & 7.4 & 0.8 & \color{blue}32.7 & 12.3 \\
    & ALOD  & SSDGMM ~\cite{choi2021active} & 5\% & 16.4 & 54.0 & 13.1 & 10.8 & 21.6 & 29.0 & 9.1 & 0 & 0 & 24.6 & 1.3 \\
    & ALOD  & \alwod &  5\%  & \color{red}37.9 & \color{red}78.9 & \color{red}38.7 & \color{red}16.4 & \color{red}58.1 & \color{red}55.0 & \color{red}11.0 & \color{red}40.2 & 9.8 & \color{red}50.1 & \color{blue}21.0\\
    \midrule
    \multirow{2}{*}{ResNet50} & WSOD
    & ${\text{D2DF2WOD}}$~\cite{wangD2DF2WOD} & 0\% & \color{blue}26.2 & \color{blue}{51.9} & \color{blue}35.5 & \color{red}18.9 & \color{blue}33.1 & \color{blue}47.4 & \color{blue}11.2 & \color{blue}{9.6} & \color{blue}{5.4} & \color{blue}{29.3} & \color{blue}{20.1}\\
   & ALOD & \alwod & 5\% & \color{red}39.3 & \color{red}79.3 & \color{red}40.0 & \color{blue}18.2 & \color{red}59.3 & \color{red}56.6 & \color{red}11.4 & \color{red}42.1 & \color{red}10.4 & \color{red}52.4 & \color{red}23.3\\
    \midrule
    Swin-T & ALOD 
    & \alwod  & 5\% & \color{red} 40.2 & \color{red}78.8 & \color{red}40.8 & \color{red}30.1 & \color{red} 58.2 & \color{red}61.2 & \color{red}12.7 & \color{red}33.2 & \color{red}12.7 & \color{red}50.9 & \color{red}23.4 \\
    \bottomrule
    \end{tabular}
    }
\end{table*}

\cref{fig:num_instance_pizza} summarizes the number of objects and images for each class on RealPizza10 dataset. In total there are 5,581 images, containing 105,781 annotated objects. In our main paper {\bf Table 1}, we summarize the detection results on three different benchmarks. Additionally, in~\cref{tab:pizza_main} we list the mean AP50 over all classes and per class AP50 values of our method and the selected baselines on RealPizza10 dataset. As shown in~\cref{tab:pizza_main}, our method significantly boosts detection performance in most classes. Based on the same backbone, our method significantly outperforms the second best D2DF2WOD~\cite{wangD2DF2WOD}. Since pepperoni is the most frequent object in RealPizza10 dataset as shown in~\cref{fig:num_instance_pizza}, at each active learning cycle, the selected images in $A^t$ always include pepperoni objects. Therefore, the detection performance of pepperoni is the best compared with other classes in active learning for object detection methods. 
The detection performance of pineapple is worse compared with other classes in active learning for object detection methods. At the initial learning stage, the AP50 of pineapple is only 1.6\% using ResNet50 backbone. There are few training images including pineapple instances, therefore the detection network can not be fully re-trained on pineapple instances. Our detection performance of broccoli is significantly better than other methods since our initial detector $M^0$ learns a good detection performance over broccoli on auxiliary images. The AP50 of broccoli is 16.6\% at the initial learning stage using ResNet50 backbone, which is higher than most baseline methods.

\begin{figure*}[hbt!]
\centering
 \includegraphics[trim={0cm 0cm 15cm 0cm},clip, scale=0.92]{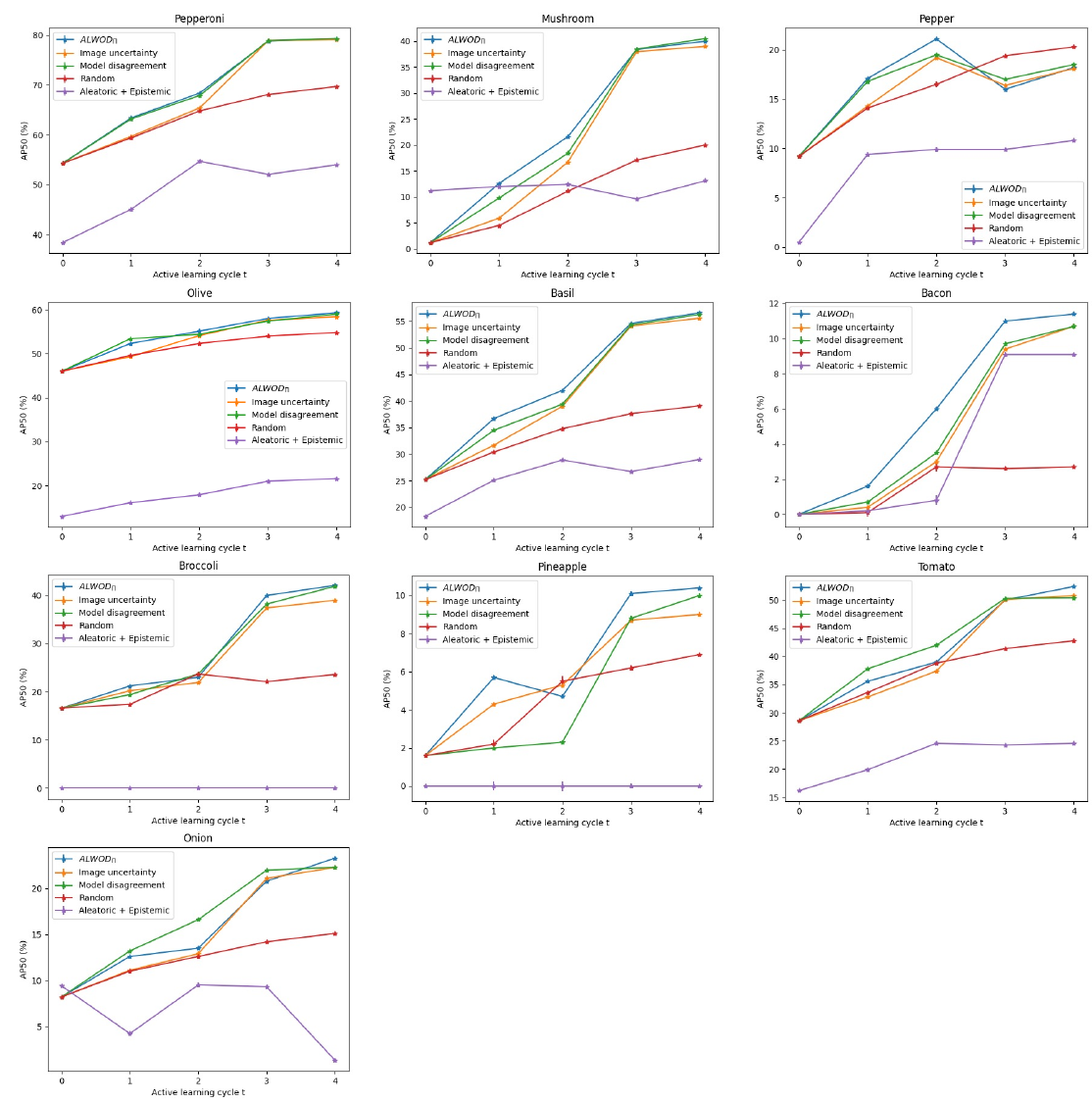}
 \caption{Per class detection performance across different active learning strategies on RealPizza10 dataset using VGG16 backbone.
 }
\label{fig:AL_perclass}
\end{figure*}

In our main paper {\bf Figure 6}, we summarize the detection performance across nine different active learning
strategies in our framework on RealPizza10 dataset. Additionally, in~\cref{fig:AL_perclass} we investigate the detection performance of each class under different acquisition functions on RealPizza10 dataset using VGG16 backbone. Since core-set~\cite{sener2017active}, loss~\cite{yoo2019learning}, and entropy-sum are worse than our proposed acquisition functions as shown in our main paper {\bf Figure 6}, and our sum strategy $\alwod_{\Sigma}$ for the final fused acquisition function is worse than the product strategy $\alwod_{\Pi}$, we do not include these active learning strategies in~\cref{fig:AL_perclass}. We include the results of SSDGMM~\cite{choi2021active} considering aleatoric and epistemic uncertainty in~\cref{fig:AL_perclass}. All the methods in~\cref{fig:AL_perclass} are
 using the same number of annotated images (5\%). \cref{fig:AL_perclass} indicates that our acquisition function outperforms other acquisition functions by a significant margin in most classes, and the detection performance of each class is improved with a higher active learning stage. At the final stage, compared with the image uncertainty score, the model disagreement score can achieve marginal improvement. At different active learning stages, the impact of model disagreement and image uncertainty is different. Fusing model disagreement and the image uncertainty in~\cref{fig:AL_perclass} suggests that these two key acquisition scores are both effective and complementary to each other. 

\section{Qualitative Evaluation}
\label{sec:qualitative}
\subsection{Selected images with proposals at each active learning cycle}
\begin{figure*}[hbt!]
\centering
 \includegraphics[trim={0cm 0cm 0cm 0},clip, scale=0.49]{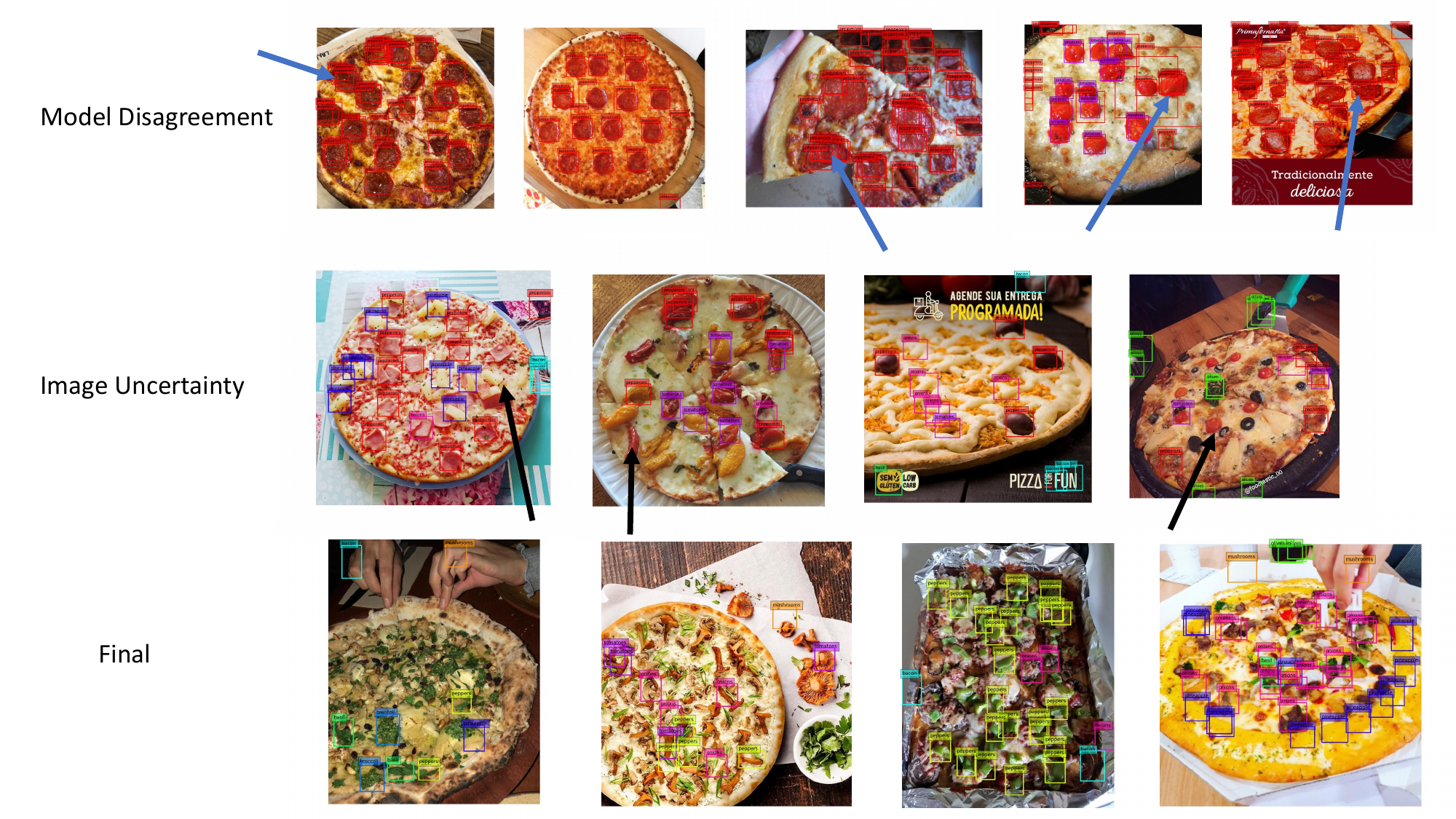}
 \caption{Example images in $A^1$ on RealPizza10 dataset. Each image includes the proposals before checked by annotators. The images selected by the model disagreement scores always include multiple redundant bounding boxes (pointed to by blue arrows). Some objects in the images selected by image uncertainty scores are missing bounding boxes (pointed to by black arrows). The final score focuses on selecting ``hard'' images, which are even challenging for the human annotators.
 }
\label{fig:step1_samples}
\end{figure*}

\begin{figure*}[hbt!]
\centering
 \includegraphics[trim={0cm 0cm 0cm 0},clip, scale=0.49]{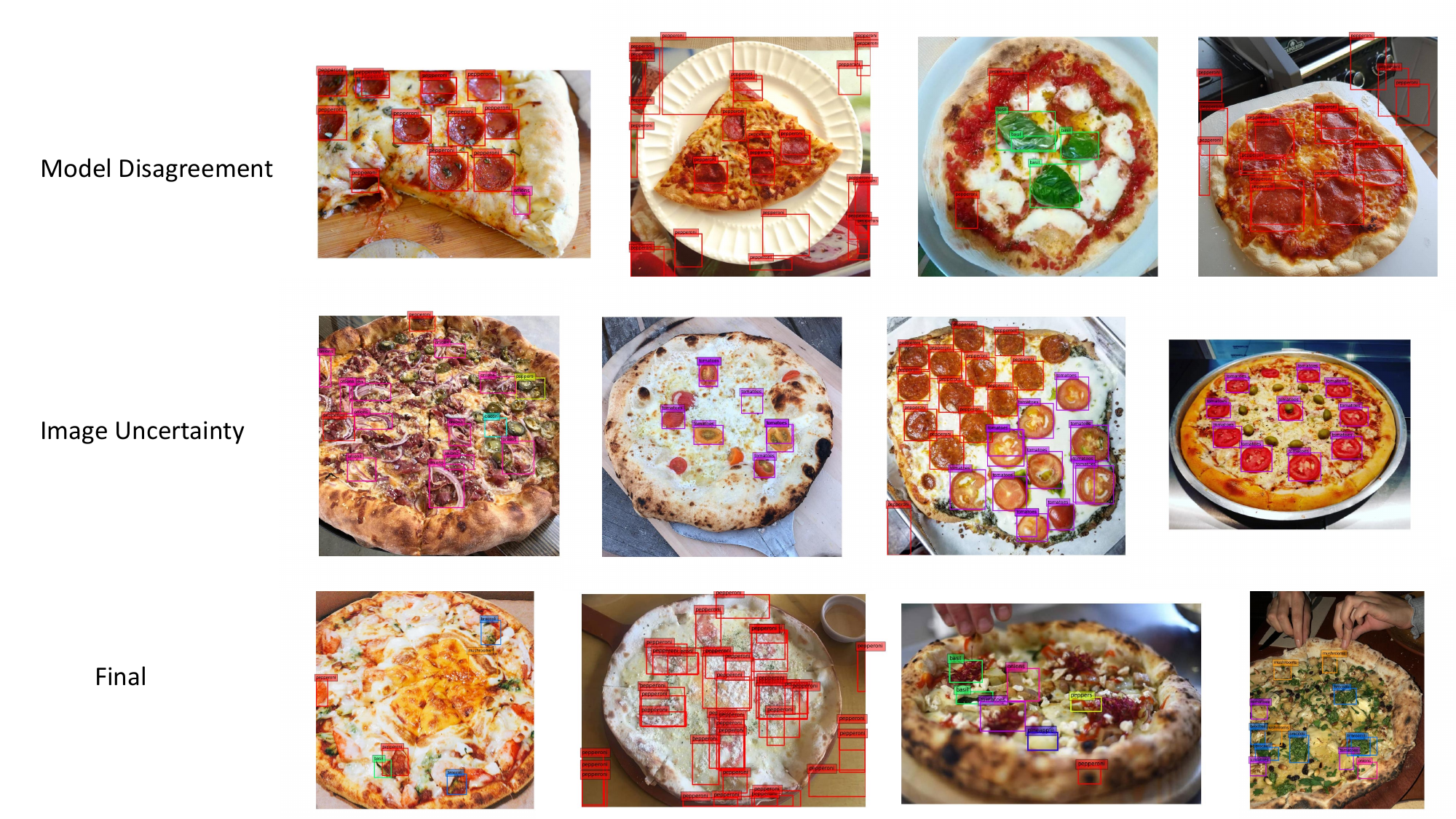}
 \caption{Example images in $A^4$ on RealPizza10 dataset. Each image includes the proposals before checked by annotators. Compared to the images in $A^1$, the images in $A^4$ are characterized by higher quality proposals.
 }
\label{fig:step4_samples}
\end{figure*}

As shown in~\cref{fig:step1_samples,fig:step4_samples},
the images selected by the model disagreement scores always include multiple redundant bounding boxes (pointed to by blue arrows). Some objects in the images selected by the image uncertainty scores are missing bounding boxes (pointed to by black arrows). The final fused score focuses on selecting ``hard'' images that even the human labelers cannot easily annotate. Also, we observe that at the fourth active learning cycle, the proposals improve in localization precision, which increases the number of precise annotations and reduces the annotation cost.

\subsection{Detection performance}
\begin{figure*}[hbt!]
\centering
 \includegraphics[trim={6cm 0cm 7cm 0},clip, scale=0.8]{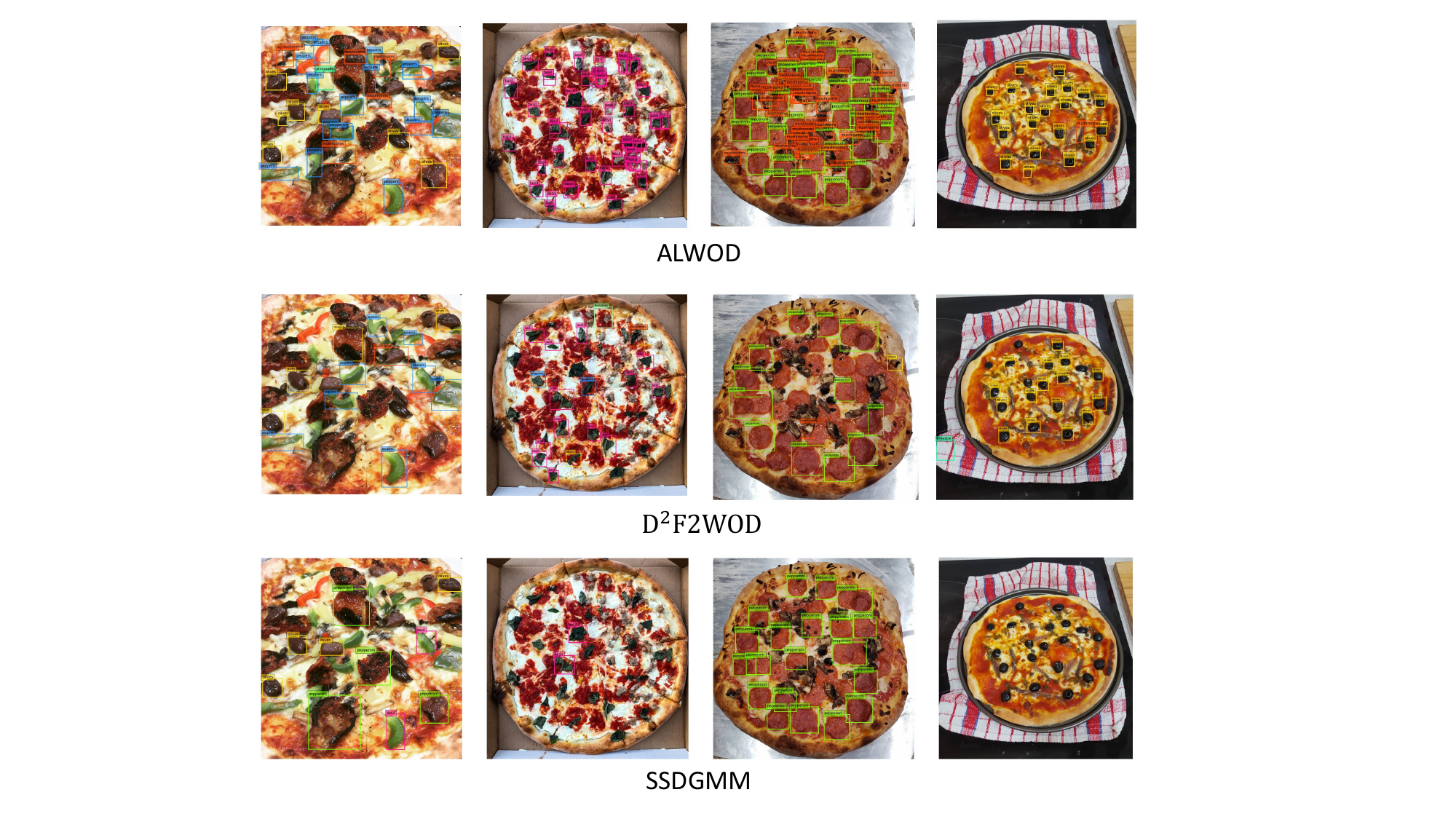}
 \caption{Examples of successful cases for \alwod vs. ${\text{D2DF2WOD}}$ vs. SSDGMM in the test set of RealPizza10 dataset. We
only show instances with scores over 0.5 to maintain visibility. It can observed that our method does not only locate most objects, but that it also
produces more accurate bounding boxes compared with other baselines. 
 }
\label{fig:final_samples}
\end{figure*}

\cref{fig:final_samples} illustrates the detection results produced by \alwod, ${\text{D2DF2WOD}}$~\cite{wangD2DF2WOD} and SSDGMM~\cite{choi2021active} on RealPizza10 dataset at the last active learning cycle. There, it can be observed that our method does not only locate most objects, but that it also
produces more accurate bounding boxes.
